\documentclass[letterpaper]{article} % DO NOT CHANGE THIS
\usepackage{aaai23}  % DO NOT CHANGE THIS
\usepackage{times}  % DO NOT CHANGE THIS
\usepackage{helvet}  % DO NOT CHANGE THIS
\usepackage{courier}  % DO NOT CHANGE THIS
\usepackage[hyphens]{url}  % DO NOT CHANGE THIS
\usepackage{graphicx} % DO NOT CHANGE THIS
\usepackage{natbib}  % DO NOT CHANGE THIS AND DO NOT ADD ANY OPTIONS TO IT
\usepackage{caption} % DO NOT CHANGE THIS AND DO NOT ADD ANY OPTIONS TO IT
\usepackage{algorithm}
\usepackage{algorithmic}

\usepackage{newfloat}
\usepackage{listings}
\DeclareCaptionStyle{ruled}{labelfont=normalfont,labelsep=colon,strut=off} % DO NOT CHANGE THIS
\floatstyle{ruled}
\newfloat{listing}{tb}{lst}{}
\floatname{listing}{Listing}
\title{DUET: Cross-modal Semantic Grounding for Contrastive Zero-shot Learning}
\author{
	Zhuo Chen\textsuperscript{\rm 1, 2, 6}, Yufeng Huang\textsuperscript{\rm 3, 6}, Jiaoyan Chen\textsuperscript{\rm 4}, Yuxia Geng\textsuperscript{\rm 1, 6}, Wen Zhang\textsuperscript{\rm 3, 6}, \\ Yin Fang\textsuperscript{\rm 1, 6}, Jeff Z. Pan\textsuperscript{\rm 5}, Huajun Chen\textsuperscript{\rm 1, 2, 6}\thanks{Corresponding Author.}
}
\affiliations{
    \textsuperscript{\rm 1}College of Computer Science and Technology, Zhejiang University\\
    \textsuperscript{\rm 2}Donghai Laboratory, Zhoushan 316021, China \\
    \textsuperscript{\rm 3}School of Software Technology, Zhejiang University \\
    \textsuperscript{\rm 4}Department of Computer Science, The University of Manchester \\
    \textsuperscript{\rm 5}School of Informatics, The University of Edinburgh \\
    \textsuperscript{\rm 6}Alibaba-Zhejiang University Joint Institute of Frontier Technologies \\
    \{zhuo.chen, huangyufeng, gengyx, wenzhang2015, fangyin, huajunsir\}@zju.edu.cn, \\
     jiaoyan.chen@manchester.ac.uk, j.z.pan@ed.ac.uk
}
\usepackage{bibentry}
\usepackage{arydshln}
\usepackage{multirow}
\usepackage{multicol}
\usepackage{bm}
\usepackage{url}
\usepackage{underscore}
\usepackage{amsfonts}
\usepackage{xcolor}
\usepackage{color, colortbl}
\usepackage{makecell}
\usepackage{enumitem}
\usepackage{amsmath}
\usepackage{booktabs}
\usepackage{amssymb}
\usepackage{soul}

\newtheorem{remark}{\noindent \textbf{Remark}}

\definecolor{mygray}{gray}{.9}
\definecolor{gray2}{gray}{.8}
\definecolor{gray3}{gray}{.7}
\definecolor{gray4}{gray}{.6}
\definecolor{gray5}{gray}{.5}
\newcommand{\wen}[1]{{\color{black}#1}}
\newcommand{\hyf}[1]{{\color{black}#1}}
\newcommand{\fy}[1]{{\color{black}#1}}
\newcommand{\gyx}[1]{{\color{black}#1}}
\newcommand{\jeff}[1]{{\color{black}#1}}

\begin{document}

\maketitle

\begin{abstract}
Zero-shot learning (ZSL) aims to predict 
\wen{unseen}
classes whose {samples} have never appeared during training. 
As annotations for class-level visual characteristics, attributes are among the {most} effective and widely used semantic information for zero-shot image classification. 
However, the current methods often fail to discriminate those subtle visual distinctions between images due to not only the lack 
 of fine-grained annotations, but also the issues of attribute imbalance and co-occurrence.  
In this paper, we present a transformer-based {end-to-end ZSL method named DUET,} 
which integrates latent semantic knowledge from the pre-trained language models (PLMs) 
{via} a self-supervised multi-modal {learning} paradigm.
Specifically, {we} \wen{(1)} develop{ed} \fy{a} cross-modal semantic grounding network to 
\fy{investigate the model's} capability of disentangling semantic attribute\fy{s} from the images;
\wen{(2)} appl{ied} an attribute-level contrastive learning strategy to further enhance the model's \fy{discrimination} on fine-grained visual characteristics against the attribute co-occurrence and imbalance; 
{ (3) {proposed} a multi-task learning policy for considering multi-model objectives.}
We find that   DUET can achieve state-of-the-art performance on three standard ZSL benchmarks and a knowledge graph equipped ZSL benchmark, and that its components are effective and its predictions are interpretable. 

\end{abstract}

\graphicspath{ {figures/} }

%% -----------------------------------------------------------------------------
%% --------------------------------Introduction---------------------------------
%% -----------------------------------------------------------------------------

\section{Introduction}
\begin{figure}[t]
  \centering
  \vspace{-1pt}
  \includegraphics[width = 0.8\linewidth]{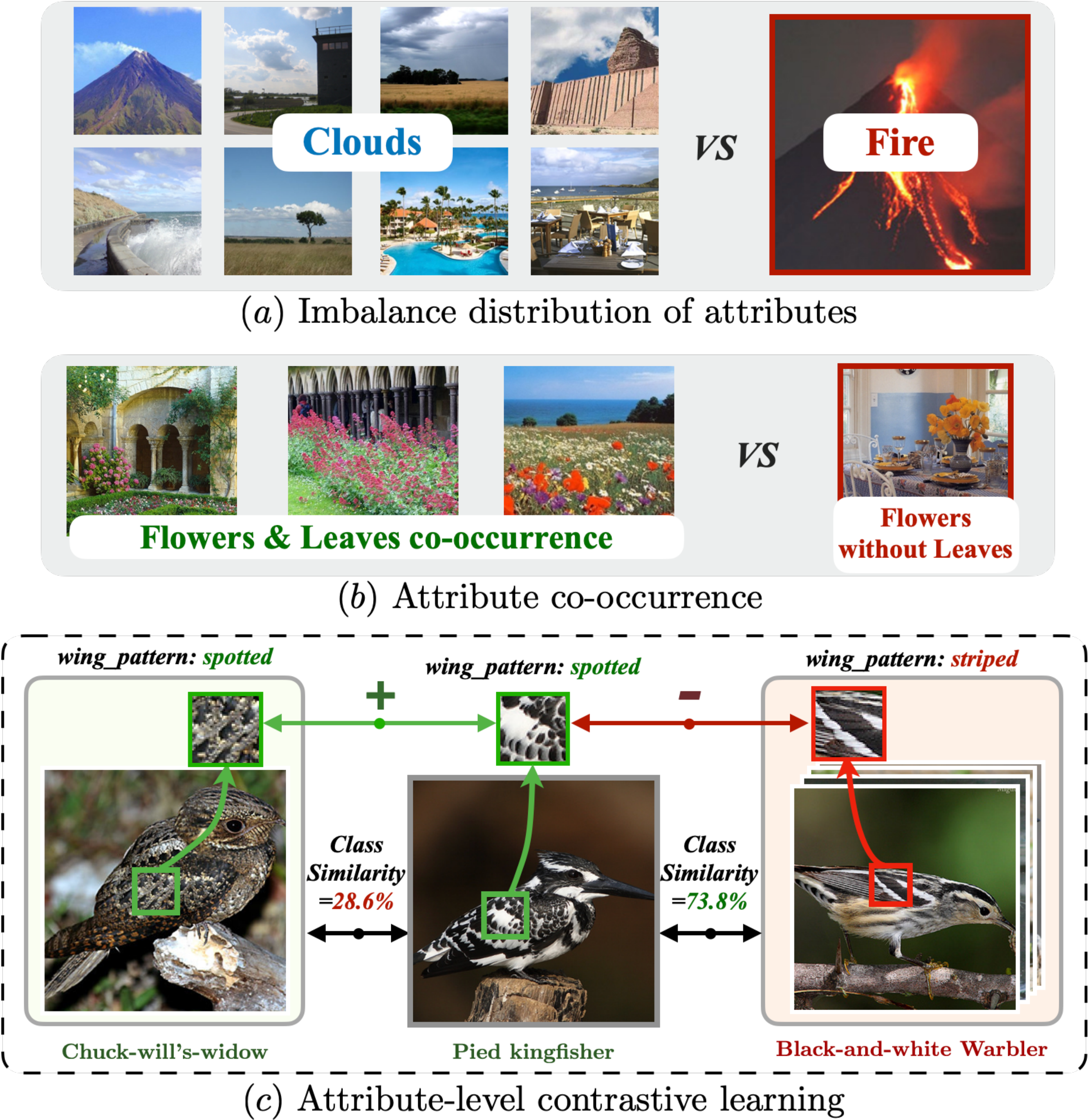}
  \caption{ 
    (a) Attribute imbalance. 
  {(b) Attribute co-occurrence.} 
  (c) {Our attribute-level contrastive learning strategy 
  which 
  chooses those distinctive classes as positive references when they are associated with one common attribute (e.g., ``spotted'')} and those similar classes as negative references 
  when they have mutually exclusive attributes (e.g., ``striped'')  toward the same aspect (e.g., ``wing_pattern'').
  }
  \label{fig:case}
  \vspace{-5pt}
\end{figure}
%  ----------task define-------------
Zero-shot learning (ZSL) aims to mimic human's inference ability to learn novel concepts based on prior experience without seeing them \fy{beforehand}.
%  ----------mapping-------------
Early embedding-based ZSL  methods  
project the input into a common vector space where  the unseen class prediction can be implemented by searching the nearest class. Generative ZSL methods create synthetic data {via the} 
side information of unseen classes, which {transforms} %could transform
ZSL into a standard supervised learning problem with less bias toward seen or unseen classes.

As annotations for image visual characteristics, attributes are among  the most popular semantic information for ZSL.
However, the attributes in real world are typically  not annotated to image regions but to a whole class. 
Recently, some attention-based ZSL methods \cite{Chen2022MSDN,Chen2021TransZero} emerge to {distinguish the discriminative regions in image classification} under the guidance of attentive attribute information.
As pointed out by \cite{DBLP:conf/cvpr/WangHLXY021}, these systems  suffer from the {\ul{\emph{imbalanced attribute distribution}}} (i.e., some attributes are highly frequent while some are rare), as well as the {\ul{\emph{attribute co-occurrence}}} which impacts attributes' discrimination capability.
{For example, in a zero-shot scene classification dataset SUN \cite{DBLP:conf/cvpr/PattersonH12}, the attributes ``trees'' and ``clouds'' are associated with 301 and 318 classes, respectively, while    ``railroad'' and ``fire'' only appear in 15 and 10 classes.}
Also, 
``flowers'' appears with ``leaves'' 39 times, but ``flowers'' alone only appears 10 times; 
% \todo{This number is 239-38 for whether grass appears together with trees}
%"grass" appears with "trees" 239 times, but "grass" alone only appears 38 times (see Appendix for detailed statistics).
% ) 
Such distribution bias may influence the model's judgment on those unseen classes which contain {rare attributes or new attribute combinations}.

\begin{figure}[htbp]
  \centering
  \vspace{-2pt}
  \includegraphics[width = 0.93\linewidth]{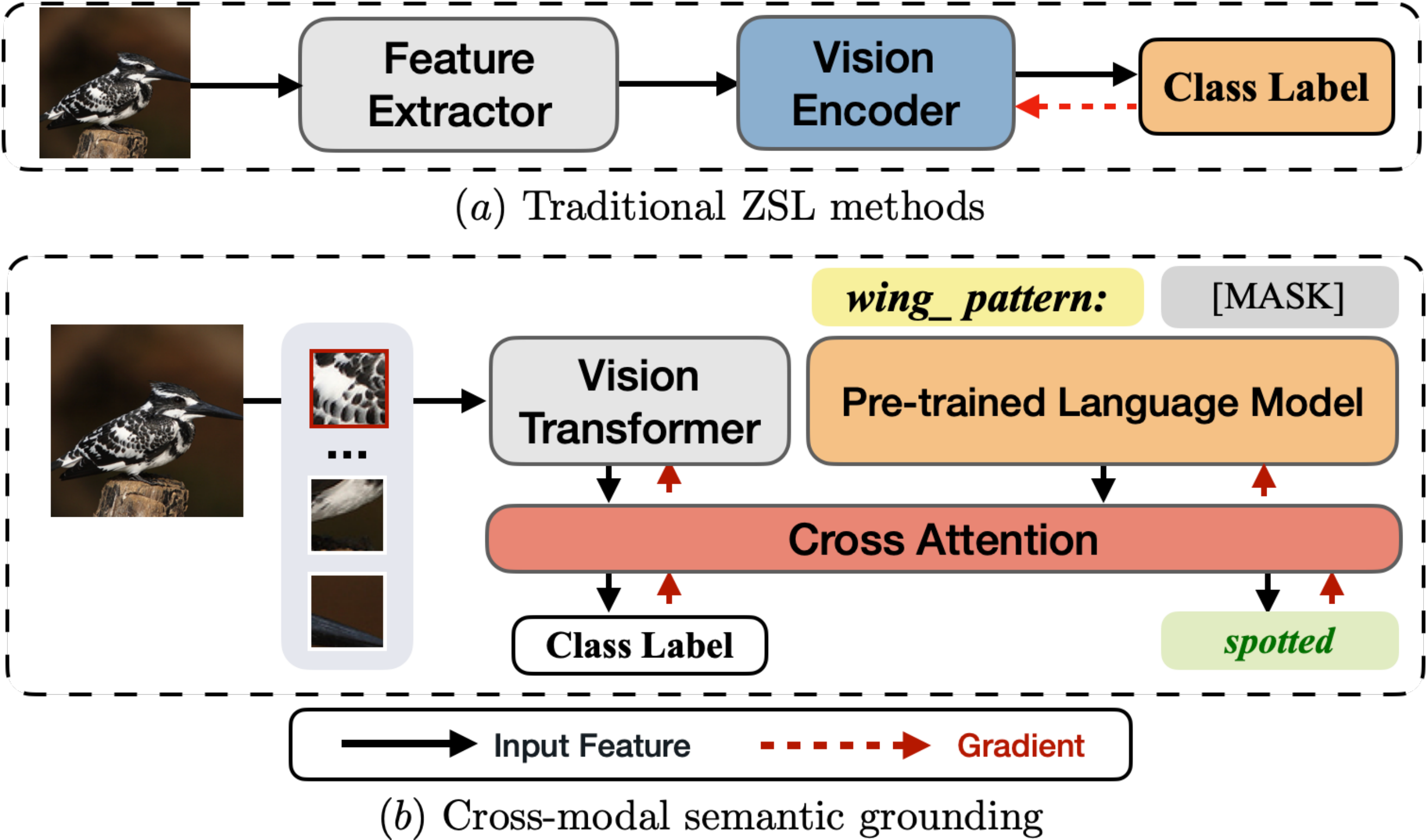}
  \vspace{-2pt}
  \caption{ (a) {The paradigm of previous ZSL methods.} 
  %{emphasize} 
  %put emphasis 
  %external class knowledge, data augmentation, or developing better vision encoders. 
  (b) {The paradigm of our method DUET 
  %We design a prompt-based cross-model semantic grounding technique, 
  which exploits the semantics of PLMs to augment the transformer-based vision encoder via reconstructing masked attributes (e.g., ``spotted'') with a cross-model attention mechanism.}}
  \label{fig:compare}
  \vspace{-12pt}
\end{figure}

%  ----------our work-------------
% Particularly,
To address these issues, %in this paper, 
we propose a novel end-to-end ZSL framework {named} \textbf{DUET} 
(Cross-mo\textbf{D}al Semantic Gro\textbf{U}nding for Contrastiv\textbf{E} Zero-sho\textbf{T} Learning).
%which emphasizes collaborative learning between the encoders from different modalities.
\fy{Unlike previous} ZSL methods in Figure \ref{fig:compare}(a) that 
%\fy{emphasized}
{emphasize utilizing more} external class knowledge, {augmenting data}, or developing better vision encoders, 
% our focus is 
\wen{we focus}
on transferring knowledge from PLMs to vision transformer encoder in a self-supervised manner, as shown in Figure \ref{fig:compare}(b), which gives \fy{the} model the ability for \ul{\emph{fine-grained semantic  grounding}} (i.e., the ability for locating relevant visual characteristics in an image given a textual attribute).

Specifically, 
% as a weakly supervised learning method, DUET consists of three parts.
we utilize a prompt-based Feature-to-Sequence Transformation (FST) proxy to 
%unify different feature forms
{transform different types of attributes into a textual sequence,} 
%into a textual format, 
which makes our model compatible to multiple {ZSL tasks with different side information.}
A Cross-modal Semantic Grounding (CSG) network is developed to leverage the semantics in a PLM  via a multi-task learning procedure, where we employ two switchable learning objective{s}: basic ZSL classification 
%(i.e., a contrastive-based ZSL objective) 
and cross-modal mask reconstruction (CMR).
Moreover, 
we propose \fy{an} attribute-level contrastive learning (ACL) \fy{mechanism} as shown in Figure \ref{fig:case}(c), where distinctive classes are selected as positive references when they are associated with one common attribute (e.g., “spotted”) and those similar classes as negative references when they have mutually exclusive attributes (e.g., “striped”) toward the same aspect (e.g., “wing\_pattern”) of the image.
This mechanism {enables} %enforces 
the model to distinguish \fy{subtle} attribute differences between closed images, and {find out the overlapped features}
%find the overlap\fy{ing} features 
between different \fy{images}.
% to enhance the vision model's ability on ZSL problem. 
The contributions can be summarized as:
 \begin{itemize}[leftmargin=*]
 \item To the best of our knowledge, DUET is the first to {investigate 
%the BERT architecture 
PLMs for zero-shot image classification. It includes a novel end-to-end multi-modal learning paradigm.}
 \item A cross-modal semantic grounding network is developed for {effective knowledge transfer from the PLM to the vision transformer encoder}.
An attribute-level contrastive learning mechanism is proposed to {address the attribute imbalance and co-occurrence issues, which further enhances \fy{the} model’s \fy{ability for distinguishing} fine-grained vision characteristics in both seen/unseen images}.
{Extensive experiments have been done on standard ZSL benchmarks equipped with attributes and one ZSL benchmark equipped with a knowledge graph. The results have verified our DUET's effectiveness and interpretation capability. }
Our code is available at \url{github.com/zjukg/DUET}.
% Our code will be available soon.
% Extensive experiments show the robustness of 
% demonstrating  flexible and adaptable
% that our proposed cross-modal semantic grounding method is valuable to handle those ZSL problems.
\end{itemize}
% -----------------------------------------------------------------------------
% -------------------------------Related Work----------------------------------
% -----------------------------------------------------------------------------
\section{Related Work}
\subsection{Zero-shot Image Classification}
\wen{The core idea of zero-shot image classification}
is to transfer semantic knowledge from seen classes to unseen classes based on their semantic information 
\cite{DBLP:conf/ijcai/ChenG0HPC21,DBLP:journals/corr/abs-2112-10006}.
% e.g., a new bird could be identified according to its common color and shape as other birds that have training images.

% \noindent 
\begin{figure*}[!htbp]
\vspace{-2pt}
  \centering
  \includegraphics[width=0.90\linewidth]{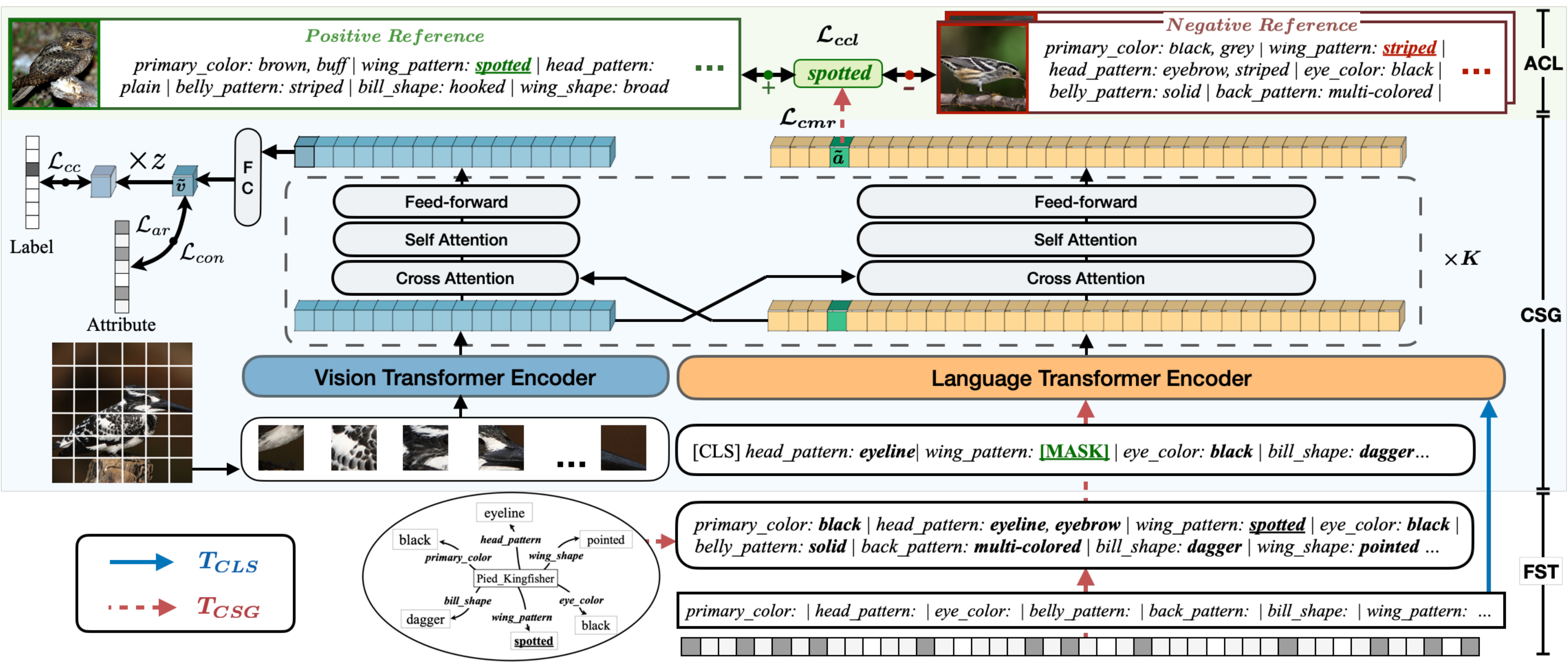}
  \vspace{-2pt}
  \caption{
  \wen{DUET consists of three parts: (1) {a} {{Feature-to-sequence transformation (FST) module}} which unifies 
%unifying 
% different feature forms 
attributes of each class 
into a textual format; 
(2) {a} {{Cross-modal semantic grounding (CSG) module}} which enables the knowledge transfer from PLM to vision transformer encoder via cross-modal mask reconstruction (CMR);
and (3) a {{Attribute-level contrastive learning (ACL) module}} which enhances the signal in CSG in a self-supervised manner.}}
  \vspace{-10pt}
  \label{fig:model}
%   \vspace{-10pt}
\end{figure*}
\textbf{Embedding-based} \wen{ZSL}
methods \cite{DBLP:conf/nips/FromeCSBDRM13} \fy{intend} to build mapping functions toward the images and/or the classes, 
% so that their vector representations after mapping are in the same space, 
and whether a class is the label of a sample can be determined by matching their vectors in the same space using similarity metrics.
% or Euclidean distance.
% It includes many ZSL methods that only map one side (either the input or the class).
However, the lack of visual samples for the unseen classes causes the bias problem and restrict their capabilities in transferring knowledge. 
% \noindent \textbf{Generative.}
To mitigate this issue, the  \textbf{Generative} ZSL methods \cite{Chen2021HSVA,DBLP:conf/iccv/0002WXPYZ021,DBLP:conf/www/GengC0PYYJC21} are introduced to use  various generative models (e.g., VAEs and GANs ) for creating synthetic data based on semantic features (e.g., simple attribute values \cite{DBLP:conf/cvpr/NaeemXTA21}, knowledge graphs \cite{PVGW2017,DBLP:conf/www/GengC0PYYJC21} or textual descriptions \cite{DBLP:conf/mir/ChenY21}). 
Those data augmentation strategies can compensate for the shortage of unseen classes and convert ZSL into a supervised classiﬁcation task. 
However, they are all complex in structure (not end-to-end) and difficult to train (owing to instability) \cite{DBLP:journals/corr/abs-2011-08641}.
\fy{The lack of} region-attribute supervision data also makes it difficult to accurately understand the corresponding relationship of different attribute-feature pairs during training. 
% \noindent \textbf{Attention-based.}
Recently, some \textbf{Attention-based} methods begin to explore the discriminative region features \fy{guided by} attentive semantic information.
% , \fy{making} the framework interpretable via the attention scores. 
Specifically, 
RGEN  \cite{DBLP:conf/eccv/Xie0ZZZYQ020} \fy{devises} the attention technique to 
construct a region graph for transferring knowledge among different classes.
GEM-ZSL \cite{DBLP:conf/cvpr/Liu00H00H21} utilize{s} gaze embedding to improve the localization of discriminative attributes.
MSDN \cite{Chen2022MSDN} incorporates mutually visual-attribute attention sub-net for semantic distillation, while TransZero \cite{Chen2021TransZero} further extend{s} MSDN via {improving the attention layers by transformers.}
However, they are still confused by the universal phenomena of
%ab,
the attribute imbalance 
and co-occurrence \cite{DBLP:conf/cvpr/ZhaoFLWWW19,DBLP:conf/cvpr/WangHLXY021}, as shown in Figure \ref{fig:case}.
Meanwhile, the scope and availability of the input knowledge limit the performance of these models on ZSL. 

\fy{In contrast to these methods}, we leverage the semantic knowledge in PLMs, and design a cross-modal semantic grounding network to encourage the model to separate those attributes from images.
Furthermore, we develop an attribute-level contrastive learning mechanism to  address the attribute imbalance and co-occurrence issues, which further enhances the model's \fy{discrimination of} different independent characteristics in a self-supervised manner.

%% -----------------------------------------------------------------------------
%% -------------------------------Methodology-----------------------------------
%% -----------------------------------------------------------------------------

\section{Methodology}
Let $\mathcal{D}_{s} = \{(x^s, y^s) | x^s \in \mathcal{X}^s, y^s \in \mathcal{Y}^s\}$ be the training set, where $x^s$ is an image with label $y^s$ attached, and $\mathcal{D}_{u} = \{(x^u, y^u) | x^u \in \mathcal{X}^u, y^u \in \mathcal{Y}^u\}$ be the unseen dataset, where 
% $\mathcal{Y}^u$ has no overlap with $\mathcal{Y}^{s}$.
\wen{$\mathcal{Y}^u$ and $\mathcal{Y}^{s}$ are disjoint.}
Each label $y$ corresponds to a class {$c \in \mathcal{C}=\mathcal{C}^{s} \cup \mathcal{C}^{u}$.}
Speciﬁcally, ZSL aims to recognize images of 
% previously 
unseen classes ($\mathcal{C}^{u}$) by transferring learned knowledge from seen classes ($\mathcal{C}^{s}$) using their {side information} (e.g., attributes).
{In this study, we assume attributes annotated to classes are given, where each attribute is sometimes associated with a real or binary value for indicating its degree.
All the attributes of a dataset are denoted as $\mathcal{A}=\{a_1,\ldots,a_\wen{|\mathcal{A}|}\}$, 
% while
\wen{and} 
the attributes of a class $c$ is denoted as $z^{c}=\left[z_{1}^{c}, \ldots, z^{c}_\wen{|\mathcal{A}|}\right]^{\top}$.}
\subsection{Feature-to-Sequence Transformation} \label{sec:FST}
\wen{This module unifies attributes of each class into a textual format. }
For ZSL datasets with binary format attributes,  we assume $a_{i} \in \mathcal{A}$ is in the attribute set $\mathcal{A}^c$ of class $c$ if $z_{i}^{c} = 1$.  
Specifically, 
we propose a prompt-based policy  to semi-serialize these discrete attributes \fy{to accommodate} the sequential input\fy{s} of PLMs, \wen{inspired by the structured tabular data pre-training \cite{DBLP:conf/acl/YinNYR20}}.
Concretely, we  cluster fine-grained attributes to define  
% $A^{\prime}$ 
\wen{$k$}
class-specific prompt  set $\mathcal{P}$ (i.e., abstract attribute set)  where $\mathcal{A}=\mathcal{P}_1\cup...\cup\mathcal{P}_{k}$.
Then, 
% for a given class, 
\wen{given a class $c$,}
we semi-serialize its attributes
with prompt (name) put ahead of each $\mathcal{P}^c$, and take special symbol ``$|$'' for prompt set separation.
%as the textual input to alleviate the impact of non-sequence.
Taking the encoded  attribute sentence {$\widehat{\mathcal{A}}^c$} for the class ``\textit{Otter}'' in AWA2 \cite{DBLP:journals/pami/XianLSA19} dataset as an example:
\begin{equation}
\small
	{\tt{..}} {\tt|} \underbrace{{\tt{color}}}_{\textrm{Prompt}} {\tt:}~  \underbrace{{\tt{brown}}}_{\textrm{Attribute}} {\tt|}  \underbrace{{\tt{has part}}}_{\textrm{Prompt}} {\tt:}~ \underbrace{{\tt{tail, flippers, ...}}}_{\textrm{Attributes}} {\tt|}	{\tt{..}}\, .
% 	\label{eq:cell_string_representation}
\end{equation}
% where $\widehat{\mathcal{A}}^c$ refers to the.
Obviously, compared to annotating large-scale fine-grained attributes for each image, it is \fy{easier} to cluster limited attribute names.
Since many ZSL datasets already have their incipient attribute divisions such as SUN \cite{DBLP:conf/cvpr/PattersonH12}, we just need to make little adjustments like removing the repeated prefixes (e.g., ``has'') and revising some ambiguous $\mathcal{P}$.
For knowledge-based ZSL (a.k.a. K-ZSL) datasets  such as AWA2-KG in OntoZSL \cite{DBLP:conf/www/GengC0PYYJC21}, given a triple
% ($ent_{c}$, $rel$, $ent_{a}$), 
\wen{($c$, $rel$, $a$), e.g., (\textit{Zebra}, \textit{hasPart}, \textit{Four_legs})}, 
we simply take the relation $rel$ as the prompt of attribute $a$. 
\subsection{Cross-modal Semantic Grounding} \label{sec:CSG}
\wen{This module enables the knowledge transfer from the PLM to the vision transformer encoder via attribute phrase masking and cross-modal mask reconstruction.}
\subsubsection{\textbf{Attribute Phrase Masking (APM).}} 
{We apply }
an
APM strategy to mask a complete attribute phrase at each step \fy{and} then urge the model to recover it.
\wen{We think \ul{\emph{discriminative  attributes with low frequency within the attribute collection are more important}}. Therefore,}
we sample the target attribute $a_t$ to be masked via a linear weighted random sampling (\textbf{LWRS}) strategy:
% \begin{equation}
$a_{t}=LWRS(\mathcal{A})$.
Given \wen{a} class $c$, 
the probability $P(a_t=a_j|\mathcal{A}^c)$ for sampling attribute $a_j$ as the $a_{t}^c$  is:
\begin{equation} \label{eq:lwrs}
P=\frac{w_{j}}{\sum_{a_{i} \in \mathcal{A}^c} w_{i}}\,, \;
w_{j}=\frac{1}{\sum_{{c}\prime \in \mathcal{C}^s} \mathbb{I}{\left[a_j \in \mathcal{A}^{{c}\prime}\right]}} \,,
\end{equation}
where $\mathbb{I}{\left[a_j \in \mathcal{A}^{{c}\prime}\right]}$ is an indicator function (i.e., it is 1 when $a_j \in \mathcal{A}^{{c}\prime}$, otherwise 0). 

Since 
the scale of non-repetitive attribute sentence $\widehat{\mathcal{A}}$ is normally much smaller than  $\mathcal{X}^{s}$ (i.e., $|\mathcal{C}^{s}| \ll |\mathcal{X}^{s}|$), 
 we propose random attributes pruning (\textbf{RAP})  over the $\mathcal{A}^c$ to remove part of the attributes (except $a_{t}$) toward a class within each training step.
%  which greatly alleviates the issue of attribute co-occurrence.
Specifically, we denote $\mathcal{A}_{rap} = RAP(r_{rap}, \mathcal{A})\,$ with hyperparameter $r_{rap}$ as the pruning ratio.
\wen{This will make }
the model to recover the attribute based on relevant visual information rather than 
% using superficial contextual cues.
\ul{\emph{trickily utilizing attribute co-occurrence}}.
\wen{Thus, the masked attribute sentence constructed based on $\mathcal{A}_{rap}$, denoted as $\widehat{\mathcal{A}}_{rap\backslash t}$,
is the input of PLM encoder.
}
\subsubsection{\textbf{Cross-modal Mask Reconstruction (CMR).}} \label{sec:CMR}
We leverage the transformer architecture to encode both the visual features and textual attributes.
Specifically, we split an image $x$ (in class $c$) into patches sequence 
 and 
{feed} them into the vision transformer encoder
with 1-D position embedding attached.
Meawhile, a learnable embedding $v_{cls}$ (marked with {\tt [CLS]}) is prepended 
% ($v^{[cls]}_L$)
whose state at the output serves as the representation of the image. 
%   and $t^{[mask]}_L = Enc_{lan}(v^{[mask]})$
Subsequently, as shown in Figure \ref{fig:model}, $K$ cross attention layers are stacked behind the parallel encoders for cross-modal information transfer. Each of them consists of one bi-directional cross-attention block, two self-attention blocks and two feed-forward blocks. A residual connection and layer normalization are added behind each block.
The keys and values \cite{DBLP:conf/nips/VaswaniSPUJGKP17} from each modality are passed as the input to other modality’s multi-headed attention blocks.
Let
$\tilde{v}$ and $\tilde{a}$ be the output representation of image $x$ and the masked target attribute, respectively.
\wen{T}he objective $\mathcal{L}_{cmr}$ for CMR is
% proposed as
\begin{equation}
\mathbb{E}_{x \sim \mathcal{X}^s}\,[ - z_{a_{t}} \sum\nolimits_{i=1}^{Len(w)} \log P(w_i\,|\,\widehat{\mathcal{A}}_{rap\backslash t}, x)] \,,
\end{equation}
where $w$ represents the token sequence of target attribute $a_{t}$ 
% within
\wen{in}
PLM's vocabulary $\mathcal{V}$, 
Specifically, $z_{a_{t}}$ is the expressive degree score for attribute  $a_{t}$ in class $c$, 
% in ground truth attribute vector $z$, 
which \ul{\emph{adaptively gives more weights to 
 those 
highly confident attributes}} (i.e., conspicuous characteristics in a class).
Moreover, we denote  
\begin{equation}
P(w_i\,|\,\widehat{\mathcal{A}}_{rap\backslash t}, x) = {\exp \left(\tilde{a}_i \cdot e_{w_i}\right)}/{\sum\nolimits_{w^\prime \in \mathcal{V}} \exp \left(\tilde{a}_i \cdot e_{w^\prime}\right)},
\end{equation}
where $e_w$ refers to the token embedding of $w$.

% Different from the complex mask logic in standard BERT \cite{DBLP:conf/naacl/DevlinCLT19}, we simply mask those 
\subsubsection{\textbf{Basic ZSL Classification.}}
Following \cite{Chen2021TransZero,DBLP:conf/nips/XuXWSA20}, we present the attribute regression loss
$\mathcal{L}_{ar}$ to encourage DUET to accurately map the image representation into  corresponding attribute embedding:
% Given a batch of $n_b$ images $x$, we denote $\mathcal{L}_{ar}$ as :
% : 
\begin{equation}
% \mathcal{L}_{ar} = \frac{1}{n_{b}} \sum\nolimits_{i=1}^{n_{b}} \| \tilde{v}_i - z^c\|^2_2 \,,
\mathcal{L}_{ar} = \mathbb{E}_{x \sim \mathcal{X}^s} \| \tilde{v} - z\|^2_2 \,,
\end{equation}
where $z$ is the class-level attribute vector for image $x$.
% - Attribute-Based Cross-Entropy Loss.
Meanwhile,
we utilize the cross-entropy loss to enforce the image to have the highest compatibility score with its corresponding class semantic vector:
\begin{equation}
\mathcal{L}_{cc}=\mathbb{E}_{x \sim \mathcal{X}^s} [-\log \frac{\exp \left(\tilde{v}\cdot  z\right)}{\sum_{\hat{c} \in \mathcal{C}^s} \exp \left(\tilde{v}\cdot  z^{\hat{c}}\right)}] \,.
\end{equation}
To further strengthen DUET's discriminative ability towards different classes with limited samples, we define a class-level supervised contrastive loss:
\begin{align}
%  \mathcal{L}_{con} &= -\frac{1}{n_{b}} \sum\nolimits_{i=1}^{n_{b}} \log f_\theta(\tilde{v}_i\,|\, s_i,  x_i)\,,
  \mathcal{L}_{con} &= \mathbb{E}_{x \sim \mathcal{X}^s} [-\log f_\theta(\tilde{v}\,|\, s,  x)]\,,
\end{align}
where $s$ is the input sentence on language side.
Specifically, let $\tilde{v}^{+}$ be the representation of positive image which has the same class label as $\tilde{v}$, those features with distinct label inside the mini-batch make up a negative samples $\mathcal{N}(\tilde{v})$. We define $f_\theta(\tilde{v}| s, x)$ as:
\begin{equation}\label{eq:cons}
f_\theta = \frac{\exp (D(\tilde{v}, \tilde{v}^{+}) / \tau)}{\exp (D(\tilde{v}, \tilde{v}^{+}) / \tau)+\sum_{\tilde{v}^{\prime} \in \mathcal{N}(\tilde{v})} \exp (D(\tilde{v}, \tilde{v}^{\prime}) / \tau)}\,,
%$}
\end{equation}
where $\tau$ is the temperature hyper-parameter. $D(\tilde{v}, \tilde{v}^{+})$ denotes the cosine similarity between $H(\tilde{v})$ and $H(\tilde{v}^{+})$, where $H(.)$ is a non-linear projection head \cite{DBLP:conf/icml/ChenK0H20}.

\subsubsection{\textbf{Multi-task Learning.} }
% \noindent \textbf{Multi-tasking Learning.} 
As the core part of cross-modal semantic grounding (CSG), 
the multi-task learning procedure \cite{DBLP:conf/nips/SenerK18,DBLP:conf/cvpr/WhiteheadWJFS21} forces the model to spread attribute information between the {vision side and the language side via a} task switching strategy. 
{Namely,} DUET conducts CMR at stage $T_{CSG}$ by accessing the visual patches and $\widehat{\mathcal{A}}_{rap}$, {and conducts} simple image classification task at stage $T_{CLS}$ without seeing {the} textual attributes. 
Specifically,  
%in order to make our model adaptive to the scenario of single-modal ZSL, 
the input sequence $s_{tmp}$ at $T_{CLS}$ {is}
%are
fixed as the prompt template to mimic the single-modal testing phase:
\begin{equation}\label{eq:template}
\small
	{\tt{..}} {\tt|} \underbrace{{\tt{color}}}_{\textrm{Prompt}} {\tt:}~ {\tt|}  \underbrace{{\tt{has part}}}_{\textrm{Prompt}} {\tt:}~  {\tt|}	 \underbrace{{\tt{pattern}}}_{\textrm{Prompt}} {\tt:}~  {\tt|} \underbrace{{\tt{shape}}}_{\textrm{Prompt}} {\tt:}~  {\tt|} {\tt{..}}\, .
% 	\label{eq:cell_string_representation}
\end{equation}
Let {${L}_{CLS}$, ${L}_{CSG}$} be the loss function of basic ZSL classification and CSG, respectively. At each step, we apply the objective $L_{CLS}$ (for $T_{CLS}$) with probability $1-\rho$, or $L_{CSG}$ (for $T_{CSG}$) with probability $\rho$:
\begin{align}
L_{CLS} &= \mathcal{L}_{zsl} + \lambda_{con} \mathcal{L}_{con}\,,\\
L_{CSG} &= \mathcal{L}_{zsl} + \lambda_{cmr} \mathcal{L}_{cmr}\,,
\end{align}
where we denote $\mathcal{L}_{zsl} = \mathcal{L}_{cc}+\lambda_{ar} \mathcal{L}_{ar}$ as the ``ZSL loss''.

\begin{table*}[!htbp]
		\centering 
% 		\small
		\caption{Results ~($\%$) of 
		{our method and the baselines. $\dag$ and $\star$ indicate generative methods and non-generative methods, respectively. }
	    The best results in baselines are marked with \underline{underline}, 
		and we highlight our results with \textbf{bold} when we achieve new SOTA.
% 		, and underline those results when they exceed all non-generative baselines.
		For CZSL, results are reported with the top-1 classification accuracy (T1). For GZSL, results are reported in terms of T1 accuracy of unseen ($U$) and seen ($S$) classes, together with their harmonic mean ($H$) where $H=(2\times S \times U)/(S+U)$.}
		\label{tab:exp}
		\vspace{-2pt}
% 		\vspace{-1mm}
		\resizebox{0.90\linewidth}{!}{
			\begin{tabular}{c|r|c|ccc|c|ccc|c|ccc}
				% \toprule
				\specialrule{.1em}{.00em}{.00em}
				&\multirow{3}{*}[-0.5ex]{\textbf{Methods}\qquad\qquad } 
				&\multicolumn{4}{c|}{\textbf{CUB}}&\multicolumn{4}{c|}{\textbf{SUN}}&\multicolumn{4}{c}{\textbf{AWA2}}\\
				% \cmidrule{3-6}\cmidrule{7-10}\cmidrule{11-14}
				\cline{3-14}
				& & \multicolumn{1}{c|}{CZSL}&\multicolumn{3}{c|}{GZSL}&\multicolumn{1}{c|}{CZSL}&\multicolumn{3}{c|}{GZSL}&\multicolumn{1}{c|}{CZSL}&\multicolumn{3}{c}{GZSL}\\
				% \cmidrule{3-6}\cmidrule{7-10}\cmidrule{11-14}
				\cline{3-14}
				\textbf{} 
				& & \rm{T1}&\rm{U} & \rm{S} & \rm{H} &\rm{T1}&\rm{U} & \rm{S} & \rm{H} &\rm{T1}&\rm{U} & \rm{S} & \rm{H} \\
				% \midrule
				\specialrule{.1em}{.00em}{.00em}
				% -------------------------
                \multirow{10}{*}{$\dag$} 
            	& TF-VAEGAN~(ECCV)~\shortcite{DBLP:conf/eccv/NarayanGKSS20} & 64.9 & 52.8 & 64.7& 58.1 & \underline{66.0} & 45.6 & \underline{40.7} & 43.0 & 72.2 & 59.8 & 75.1 & 66.6 \\
				& Composer~(NeurIPS)~\shortcite{DBLP:conf/nips/HuynhE20}&69.4&56.4&63.8&59.9&62.6& \underline{55.1}&22.0& 31.4&71.5&62.1&77.3&68.8\\
            	& CE-GZSL~(CVPR)~\shortcite{DBLP:conf/cvpr/HanF0021} & 77.5 & 63.1 & 66.8 & 65.3 & 63.3 & 48.8 & 38.6 & 43.1 & 70.4 & 63.1 & 78.6 & 70.0 \\
            	& GCM-CF~(CVPR)~\shortcite{DBLP:conf/cvpr/YueWS0Z21}&--&61.0&59.7&60.3&--& 47.9&37.8& 42.2&--& 60.4&75.1&67.0\\
				& FREE~(ICCV)~\shortcite{DBLP:conf/iccv/0002WXPYZ021}&--&55.7&59.9&57.7&--& 47.4&37.2& 41.7&--& 60.4&75.4&67.1\\
				& HSVA~(NeurIPS)~\shortcite{Chen2021HSVA}&62.8&52.7&58.3&55.3&63.8&48.6&39.0&\underline{43.3}&--&59.3&76.6&66.8\\
            	& AGZSL~(ICLR)~\shortcite{DBLP:conf/iclr/ChouLL21} & 57.2 & 41.4 & 49.7 & 45.2 & 63.3 & 29.9 & 40.2 & 34.3 &\underline{73.8} & 65.1 & 78.9 & 71.3 \\
            % 	\midrule
                \specialrule{.1em}{.00em}{.00em}
            	\multirow{8}{*}{$\star$} 
				& APN~(NeurIPS)~\shortcite{DBLP:conf/nips/XuXWSA20}&72.0&65.3& 69.3&67.2&61.6& 41.9&34.0&37.6&68.4&57.1&72.4&63.9\\
				& DVBE~(CVPR)~\shortcite{DBLP:conf/cvpr/MinYXWZZ20}&--&53.2&60.2&56.5&--&45.0&37.2&40.7&--&63.6&70.8&67.0\\ 
				& DAZLE~(CVPR)~\shortcite{DBLP:conf/cvpr/HuynhE20}&66.0&56.7&59.6&58.1&59.4&52.3&24.3&33.2&67.9&60.3&75.7&67.1\\
				& RGEN~(ECCV)~\shortcite{DBLP:conf/eccv/Xie0ZZZYQ020} & 76.1 & 60.0 & \underline{73.5} & 66.1 & 63.8 & 44.0 & 31.7 & 36.8 & 73.6 & \underline{67.1} & 76.5 & 71.5 \\
                & GEM-ZSL~(CVPR)~\shortcite{DBLP:conf/cvpr/Liu00H00H21} & \underline{77.8} & 64.8 & 69.3 & 67.2 & 62.8 & 38.1 & 35.7 & 36.9 & 67.3 & 64.8 & 77.5 & 70.6 \\
            	& MSDN~(CVPR)~\shortcite{Chen2022MSDN} & 76.1 & 68.7 & 67.5 & 68.1 & 65.8 & 52.2 & 34.2 & 41.3 & 70.1 & 62.0 & 74.5 & 67.7 \\
				& TransZero~(AAAI)~\shortcite{Chen2021TransZero} &76.8&\underline{69.3}&68.3&\underline{68.8}&65.6&52.6&33.4&40.8&70.1&61.3&\underline{82.3}&70.2\\
				% \cdashline{2-13}[4pt/1pt]
				% \midrule
				\cmidrule{2-14}
				% \hline{2-14}
				& {\textbf{DUET}}{~\textbf{(Ours)}}    &72.3&62.9&72.8&67.5&64.4&45.7&\textbf{45.8}&\textbf{45.8}&69.9&63.7&\textbf{84.7}&\textbf{72.7}\\
				% \bottomrule	
				\specialrule{.1em}{.00em}{.00em}
		\end{tabular} 
		}
		\vspace{-7pt}
		\label{table:sota}
	\end{table*}

\subsection{Attribute-level Contrastive Learning} \label{sec:ACL}
To further strengthen the model's sensitivity on subtle  visual differences against the attribute co-occurrences,
we introduce an attribute-level contrastive learning (ACL) module with the adaptive loss function:
\begin{align}\label{eq:pool}
 \mathcal{L}_{acl} &= \mathbb{E}_{x \sim \mathcal{X}^s} [- Min(z_{{a}}, z_{{a}^+})\log f_\phi(\tilde{a}\,|\, s, x)]\,,
\end{align}
where $f_\phi(\tilde{a}\,|\, s, x)$ follows the base formulation of Eq. (\ref{eq:cons}),  but there are $3$ main differences between $f_\phi$ and $f_\theta$:

\textit{(i)} \textit{Target Object and Stage}. $f_\phi$ targets at the mean-pooling representation of $\tilde{a}$  on language side of stage $T_{CSG}$, where the input sentence $s$ is $\widehat{\mathcal{A}}_{rap\backslash t}$.
While $f_\theta$ targets at the feature ($\tilde{v}$) on vision side, which is applied at stage $T_{CLS}$ with a fixed prompt template (\ref{eq:template}) as the $s_{tmp}$.

\textit{(ii)} \textit{Sampling Strategy}. 
For class-level $f_\theta$, we simply pick those images, which share the same class label with original sample as positive,  and then define the rest as in-batch negative.
While for attribute-level $f_\phi$, we design {an}
%a
attribute-based sampling strategy:
Given a class $c$ and its target attribute $a^c_{t}$, we assume $a^{c-}_{t}$ as the negative attribute from seen class $c^-$, and $a^{c+}_{t}$ as the positive attribute from seen class $c^+$. 
We claim the precondition as:
\begin{align}
    c\neq c^+\neq c^-, \;
    a^c_{t} = a^{c+}_{t} \neq a^{c-}_{t}\,, 
    \\
    a^c_{t}\,, \,a^{c-}_{t} \in \mathcal{P}_{t},\;\; 
    a^{c-}_{t} \notin \mathcal{P}_{t}^{c}, \;
    a^{c}_{t} \notin \mathcal{P}_{t}^{c-}\, ,
\end{align}
where $\mathcal{P}_{t}$ is the original {{class-agnostic}} prompt set that $a_{t}$ belongs to, and $\mathcal{P}_{t}^{c}$, $\mathcal{P}_{t}^{c-}$ is the {{class-specific}} prompt set in class $c$, $c^-$.
All $c^+$, $c^-$ that satisfies this precondition make up the candidate class set $\mathcal{C}^+$ and $\mathcal{C}^-$, respectively.

\textit{(iii)} \textit{Sampling Probability}. 
We employ a heuristic process to let the model 
%prefer to 
select those $c^+$ whose $\mathcal{A}^{c+}$ are more \ul{\emph{inconsistent}}, and  $c^-$ whose $\mathcal{A}^{c-}$ are more \ul{\emph{similar}}, compared with $\mathcal{A}^c$. 
Then, we non-repetitively choose instances (i.e. $x^{c-}$ and $x^{c+}$) from these classes, %then
{and} encode them by DUET to get the final $\tilde{a}^-$ and $\tilde{a}^+$.
{Note that} $a^{c-}_{t}$ and $a^{c+}_{t}$ {are not} masked to accelerate the convergence. 

Finally, we stack this pluggable ACL module into CSG:
\begin{align}
 L_{CSG} \longleftarrow L_{CSG} + \mathcal{L}_{acl}\,.
 \vspace{-2pt}
\end{align}
\begin{remark}
\vspace{-7pt}
{Considering the example in Figure \ref{fig:case},} we assume ``Pied Kingfisher'' as the original bird class ($c$) with  target attribute ``spotted'' ($a_{t}$) in {the} prompt set ``wing pattern'' ($\mathcal{P}_{t}^{c}$).  
We are likely to sample  ``Chuck will's widow'' as the positive class $c^+$ which {contains} spotted wing pattern, but has a low class similarity (28.6\% after normlization) compared with ``Pied Kingfisher''. Besides, we prefer to sample ``Black-and-white Warbler'' as the negative class $c^-$ whose wing pattern is striped ({not} ``spotted'') but the class characteristic is pretty closed (73.8\%) to ``Pied Kingfisher''.
%\vspace{-2pt}
\end{remark}

\subsection{Zero-Shot Prediction}\label{sec:ZSP}
After training DUET, we use the learned cosine metric space for zero-shot recognition.
There are two evaluation settings: \fy{conventional ZSL (CZSL), which classifies the testing samples with candidate classes from $\mathcal{C}^u$;
generalized ZSL (GZSL), which} 
extends the candidate classes to $\mathcal{C}^s \cup \mathcal{C}^u$. 
Specifically, we take the prompt template $s_{tmp}$ 
together with an test image $x$ as the input.
Following \cite{DBLP:conf/cvpr/Liu00H00H21,Chen2022MSDN}, we 
predict the label $c^*$ via:
\begin{equation}
c^{*}=\underset{c \in \mathcal{C}^u / \mathcal{C}}{\arg \max } \left(\tilde{v} \cdot z^c \right) - \gamma \mathbb{I}\left[c \in \mathcal{C}^{s}\right]\,,
\end{equation}
where $\mathbb{I}=1$ if $c$ is a seen class and 0 otherwise. $\gamma$ is the calibration factor tuned on a held-out validation set, and  $\mathcal{C}^u / \mathcal{C}$ corresponds to the CZSL/GZSL setting respectively.

%% -----------------------------------------------------------------------------
%% --------------------------------Experiments----------------------------------
%% -----------------------------------------------------------------------------

\section{Experiments}
\subsection{Datasets}

{We select three standard attribute equipped ZSL benchmarks \textbf{AWA2} \cite{DBLP:journals/pami/XianLSA19}, \textbf{CUB} \cite{welinder2010caltech}, \textbf{SUN} \cite{DBLP:conf/cvpr/PattersonH12} with their splits proposed in \cite{DBLP:journals/pami/XianLSA19}, as well as a knowledge graph equipped benchmark \textbf{AWA2-KG} which has the same split as AWA2 but includes semantic information about hierarchical classes and attributes, for evaluation. 
In AWA2-KG,}
we assume \fy{that} the class $c$ \fy{has} the attribute $a_i$ when the length of the shortest same-direction relation path between \fy{them} in KG is $h$, where $h$ is a hyperparameter. 
For example, given two triples (\textit{Zebra}, \textit{hasPart}, \textit{Four_legs}) and (\textit{Four_legs}, \textit{subClassOf}, \textit{Leg}), the attribute of class ``Zebra'' is ``Four\_leg'' when $h$$=$1 and ``Leg'' when $h$$=$2.
{\fy{Since we observe that the attribute $a$ become more coarse-grained when they are far away from the class $c$ in KG, we simply define $h$ as $1$.}}

Besides, there are also some other 
% popular 
ZSL datasets like aPY \cite{DBLP:conf/cvpr/FarhadiEHF09} and AWA1 \cite{DBLP:journals/pami/LampertNH14}.  \jeff{We} do not evaluate on them since most of their unseen classes are leakage \cite{DBLP:journals/pami/XianLSA19}  in ImageNet-1K \cite{DBLP:conf/cvpr/DengDSLL009}, which is a common image datasets for normal vision encoder pre-training. 

%% --------------------------------Experimental Settings----------------------------------
\subsection{Experimental Settings}
Unlike previous ZSL studies which pre-extract the image features using a pre-trained CNN model e.g., ResNet \cite{DBLP:conf/cvpr/HeZRS16}, we take as input the raw images and apply vision transformer to 
interact with the PLM for knowledge transfer.
For those coefficients in AWA2, we set $\lambda_{ar}$ to \hyf{0.01}, $\lambda_{con}$ to \hyf{0.05}, $\lambda_{cmr}$ to \hyf{1}, $\lambda_{acl}$ to 0.01,  $r_{rap}$ to 0.5, $\rho$ to 0.4 and $\gamma$ to 0.8. 
{We report the class-averaged (macro) accuracy as the basic metric, following the current literature \cite{DBLP:conf/nips/XuXWSA20,Chen2021TransZero}.}

\subsection{Overall Results}
%% -------------------------------- overall result ----------------------------------
% \noindent \textbf{Standard ZSL Datasets.}
\subsubsection{\textbf{Standard ZSL Datasets.}}
We compare our method with \textbf{14} representative {or state-of-the-art (SOTA) methods proposed in recent three years.}
{These baselines are divided into two \gyx{categories}:}
%(
non-generative \cite{DBLP:conf/nips/XuXWSA20,DBLP:conf/cvpr/MinYXWZZ20,DBLP:conf/cvpr/HuynhE20,DBLP:conf/eccv/Xie0ZZZYQ020,DBLP:conf/cvpr/Liu00H00H21,Chen2022MSDN,Chen2021TransZero} and generative \cite{DBLP:conf/eccv/NarayanGKSS20,DBLP:conf/nips/HuynhE20,DBLP:conf/cvpr/HanF0021,DBLP:conf/cvpr/YueWS0Z21,DBLP:conf/iccv/0002WXPYZ021,Chen2021HSVA,DBLP:conf/iclr/ChouLL21}.
%). 
All those non-generative methods are attention-based except {for} DVBE \cite{DBLP:conf/cvpr/MinYXWZZ20}.

\textbf{SUN} contains more than 700 scene classes but each class has {only 10-20 images instances, where the attribute} imbalance and co-occurrence problem are universal.
We find that DUET achieves the best accuracy (45.8$\%$) \fy{on} $H$ with a large margin (2.5$\%$) compared with those SOTA methods, and surpass MSDN by 4.5$\%$ on $H$, which is the SOTA no-generative methods on SUN.
{On \textbf{AWA2}, DUET gains 1.2$\%$ improvements over the SOTA performance}, and outperforms the transformer-based method TransZero  \fy{on} all {the} GZSL metrics.
{On \textbf{CUB},} DUET achieves competitive performance\fy{, surpassing all generative methods on $H$, {except for the attention-based methods} TransZero and MSDN.}
We own this to the fact that the prompts in CUB are mostly region-related (e.g., ````breast\_color'' and ``wing\_color''),  but DUET simply 
%attach
{attaches} the image patches with sequential 1-dimensional positional embedding as the input, {making it hard to capture} the fine-grained positional relationship.
% {using} only 7,057 training images.
Instead, TransZero {takes}
%take
2D  center coordinates to construct learnable relative region \fy{geometric} embeddings for feature augmentation, which gets accurate position representations and {helps
the model achieve good performance}.  
{Notably, when it does not use} feature augmentation from relative geometry relationships, the $H$ on CUB dramatically decreases to 66.5$\%$ \cite{Chen2021TransZero}.

{It is worth mentioning that DUET also achieves great performance on  seen classes ($S$) on all three datasets, \fy{outperforming} all baselines on SUN and AWA2 by at least 5.1$\%$ and 2.4$\%$ respectively.
This \fy{proves} that DUET well preserves the predictive ability {on} seen class\fy{es} in addressing unseen classes.}

\subsubsection{\textbf{K-ZSL Dataset.}}
\begin{figure}[htbp]
  \centering
  \vspace{-5pt}
  \includegraphics[trim=10 0 0 0, width = 1\linewidth]{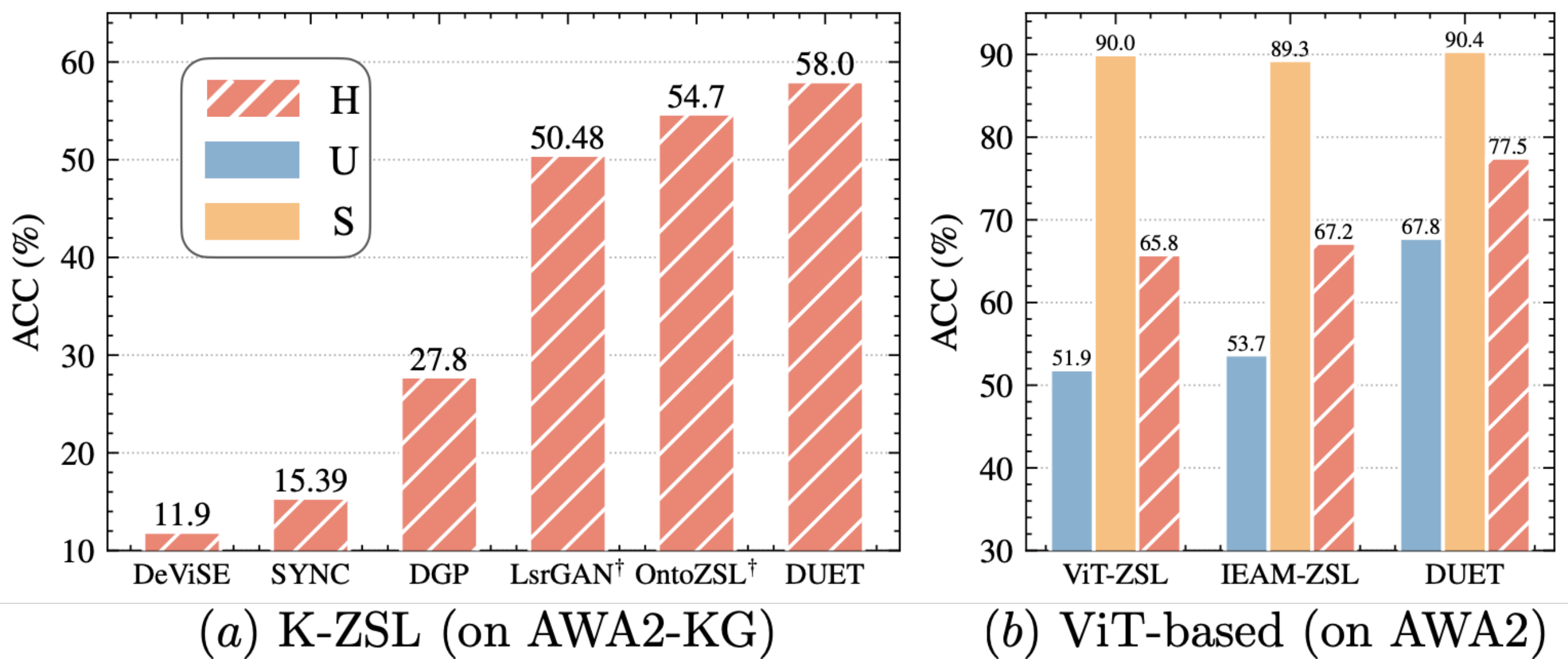}
  \caption{ (a) Results ~($\%$) on AWA2-KG from OntoZSL. The attribute values in this dataset are all represent in 0/1 binary form. We marks those generative methods with ``$\dag$''. (b) Results ~($\%$) on AWA2 with ViT-base as the vision encoder.}
  \label{fig:kgvit}
  \vspace{-7pt}
\end{figure}

We evaluate on AWA2-KG \cite{DBLP:conf/www/GengC0PYYJC21} to validate DUET's  flexibility on various ZSL attribute formats.
Specifically,
we pick the KG from \cite{geng2021k} as the knowledge resource, and compare with  baselines
%K-ZSL benchmarks
including DeViSE \cite{DBLP:conf/nips/FromeCSBDRM13}, SYNC \cite{changpinyo2016synthesized}, DGP \cite{DBLP:conf/cvpr/KampffmeyerCLWZ19}, LsrGAN$^\dag$ \cite{vyas2020leveraging}. 
We abandon the real-value attributes for {fairness}, and follow \cite{DBLP:conf/www/GengC0PYYJC21,DBLP:journals/corr/abs-2206-03739,DBLP:conf/semweb/0007CGPYC21} to take the KG embedding \cite{DBLP:conf/nips/BordesUGWY13} for entity class representation toward $\mathcal{L}_{ar}$ and $\mathcal{L}_{cc}$.
As shown in Figure \ref{fig:kgvit}(a), DUET achieves higher performance \fy{among} all other methods. In particular, {it achieves \fy{a} 30.2$\%$ improvement \fy{on} metric $H$} compared to the non-generative method DGP.
\subsubsection{\textbf{{ViT-based DUET.}}}
To get further insights into our model, we report the results of DUET with ViT-base \cite{DBLP:conf/iclr/DosovitskiyB0WZ21} as the vision encoder.
Remarkably, since the released ViT-base is pre-trained on ImageNet-21K \fy{which} may contain unseen objects,
we only select 2 recent ViT-based ZSL methods, ViT-ZSL~\cite{DBLP:journals/corr/abs-2108-00045} and IEAM-ZSL~\cite{DBLP:conf/eccv/NarayanGKSS20}, for comparison.
As shown in Figure \ref{fig:kgvit}(b), DUET surpasses these two methods by a large margin (14.1$\%$ improvement \fy{on} $U$ and 10.3$\%$ improvement \fy{on} $H$) and also exceeds our SOTA performance ($H$) by 4.8$\%$. 
%which demonstrates {DUET's compatibility towards different vision encoders.}
This supports that our DUET greatly ameliorates the ZSL ability where the original vision transformer is poor.
We believe that the performance will be further improved \fy{by plugging in a better vision transformer encoder.}

\subsection{Ablation Studies}
\begin{table}[!htbp]
		\centering  
% 		\small
        \vspace{-5pt}
		\caption{Results ~($\%$) of ablation studies on AWA2 \fy{by GZSL}. \fy{The metric is harmonic mean ($H$) accuracy}. $\vartriangle$ indicates the performance drop compared with our full model. }
		\vspace{-3pt}
% 		\vspace{-1mm}
 		\resizebox{0.65\linewidth}{!}{
			\begin{tabular}{l|cc}
% \toprule
\specialrule{.1em}{.00em}{.00em}
Methods  
&  \rm{H} &  $\vartriangle$\\
\specialrule{.1em}{.00em}{.00em}
% -------------------------
Only $ENC_{vis}$ & 64.1  & 8.6\color{blue}{$\downarrow$} \\
\cdashline{1-3}[4pt/1pt]
$1)$  CSG$_{freeze ~ENC_{lan}}$    & 66.5 & 6.2\color{blue}{$\downarrow$}  \\
$2)$  CSG$_{w/~only~Prompt}$   & 61.7 & 11.0\color{blue}{$\downarrow$}  \\
$3)$  CSG$_{w/o~Prompt}$   & 64.9 & 7.8\color{blue}{$\downarrow$} \\
$4)$  CSG$_{w/o~LWRS}$   & 66.9 & 5.8\color{blue}{$\downarrow$}  \\
$5)$  CSG$_{w/o~RAP}$   & 67.4 & 5.3\color{blue}{$\downarrow$}  \\
$6)$  CSG$_{w/o~\mathcal{L}_{con}}$    & 68.4 & 4.3\color{blue}{$\downarrow$}\\
$7)$  CSG   & 69.2 & 3.5\color{blue}{$\downarrow$}  \\
% -------------------------
\specialrule{.1em}{.10em}{.00em}
{\textbf{DUET}}~{\textbf{(Full model)}} & \textbf{72.7} & -\\
\specialrule{.1em}{.00em}{.00em}	
		\end{tabular} 
 		}
		\label{tab:ablation}
		\vspace{-7pt}
\end{table}
% \noindent \textbf{Component Analysis.}
\subsubsection{\textbf{Component Analysis.}}
We evaluate various stripped-down versions of our proposed model to compare the  ($H$) performance gain brought by different components on AWA2.
Concretely, we observe that the performance drop{s} sharply when 
\textbf{(1)} freezing the language transformer encoder $ENC_{lan}$. Although it
  \fy{can} reduce the overall learnable parameters, it makes the model hard{er} to understand the special relationship among prompts, textual attributes, and visual features.
From the results of taking \textbf{(2)} only the prompt and \textbf{(3)} only concatenating attribute as sequence input without the prompt, we observe that employing our FST strategy for semi-serializing attributes
indeed benefit{s} our model with 4.3$\%$ improvement.
We also exploit the influence of \textbf{(4)} {randomly} masking attributes, \textbf{(5)}  not conducting attribute pruning, which leads to 2.3$\%$, 1.8$\%$ falls compared with \textbf{(7)} applying the full CSG, proofing the necessity of both sampling target attribute with adaptive weight and pruning part of the attribute.
Besides,  \textbf{(6)} abandoning class-level contrastive learning leads to 0.8$\%$ decrease.
%w
We own this to the fact that contrastive learning can help model learn better visual representations by narrowing  the distance within a class in the latent space.
%which help the prediction.
Most importantly, our pluggable ACL module {further}
%futher
boosts the  performance by 3.5$\%$ based on CSG, which illustrates that both of these modules are {beneficial.}

\begin{figure}[htbp]
  \centering
   \vspace{-5pt}
  \includegraphics[trim=30 0 0 0, width = 1.0\linewidth]{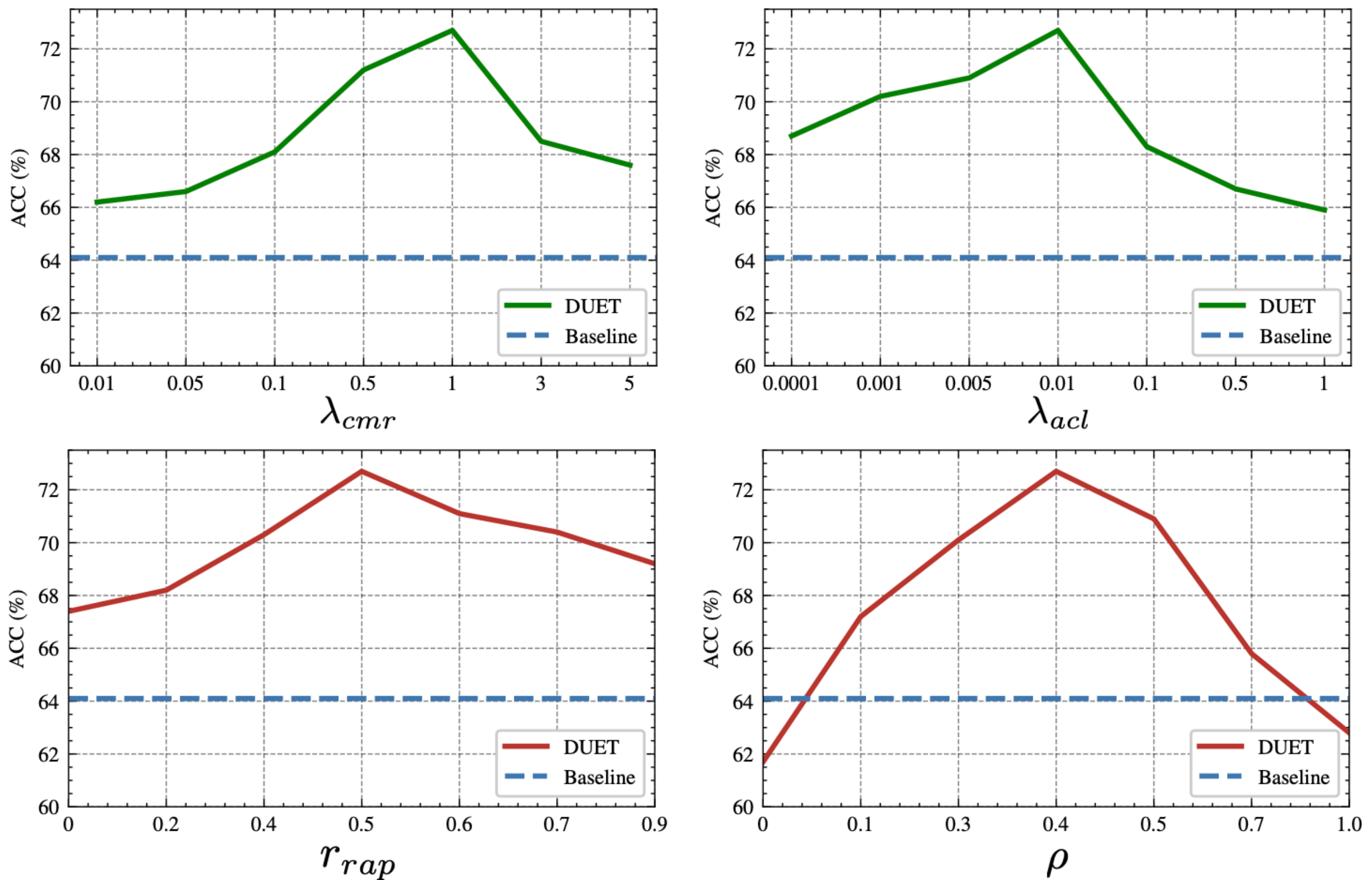}
  \vspace{-5pt}
  \caption{A parameter analysis of coefficients $\lambda_{cmr}$, $\lambda_{acl}$, $r_{rap}$ (the pruning ratio for $RAP(.)$), $\rho$ (the probability for employing $L_{CSG}$). Baseline denotes the pure vision trasnformer encoder.
  {\rm{H} accuracy on AWA2 is reported.}
  %All ablation studies are test on AWA2 by  harmonic mean.
  }
  \label{fig:parameter}
  \vspace{-10pt}
\end{figure}
\subsubsection{\textbf{Hyperparameter Analysis.}}
% We analyze the effect of key hyperparameters within our model, and the results are shown.
By comparing DUET's performance in Figure \ref{fig:parameter}, 
%we can draw the following conclusion: 
we conclude that:
\textit{(i)} 
The performance decrease when $\lambda_{cmr}$ and $\lambda_{acl}$ are too large, since the weak signal from the self-supervised objective{s} (i.e., $\mathcal{L}_{cmr}$ and $\mathcal{L}_{acl}$) will gradually overwhelm the signal from supervised class label (i.e., $\mathcal{L}_{ar}$ and $\mathcal{L}_{cc}$).
\textit{(ii)} 
%  $\rho$ determines the probability for employing $L_{CSG}$.
When $\rho$ is close to 1 or 0, the protocol all drops below the baseline.
This is because the model turns into a single-modal task without multi-task learning when $\rho=0$. While when $\rho=1$, DUET is forced to classify the image with attribute attached \fy{throughout} the training, \fy{leading to} model's poor generalization capability at test stage.
\textit{(iii)} 
Furthermore, we try a wide range of $r_{rap}$, i.e. $r_{rap}$ = \{0, 0.2, 0.4, 0.5, 0.6, 0.7, 0.8\}, and \fy{find that} DUET \fy{works best} when $r_{rap}$ is set to 0.5.

%% -----------------------------------------------------------------------------
%% --------------------------------Experiments----------------------------------
%% -----------------------------------------------------------------------------
% \vspace{-3pt}
\subsection{Interpretation
%Interpretable
}
\begin{figure}[htbp]
  \centering
  \includegraphics[width=0.99\linewidth]{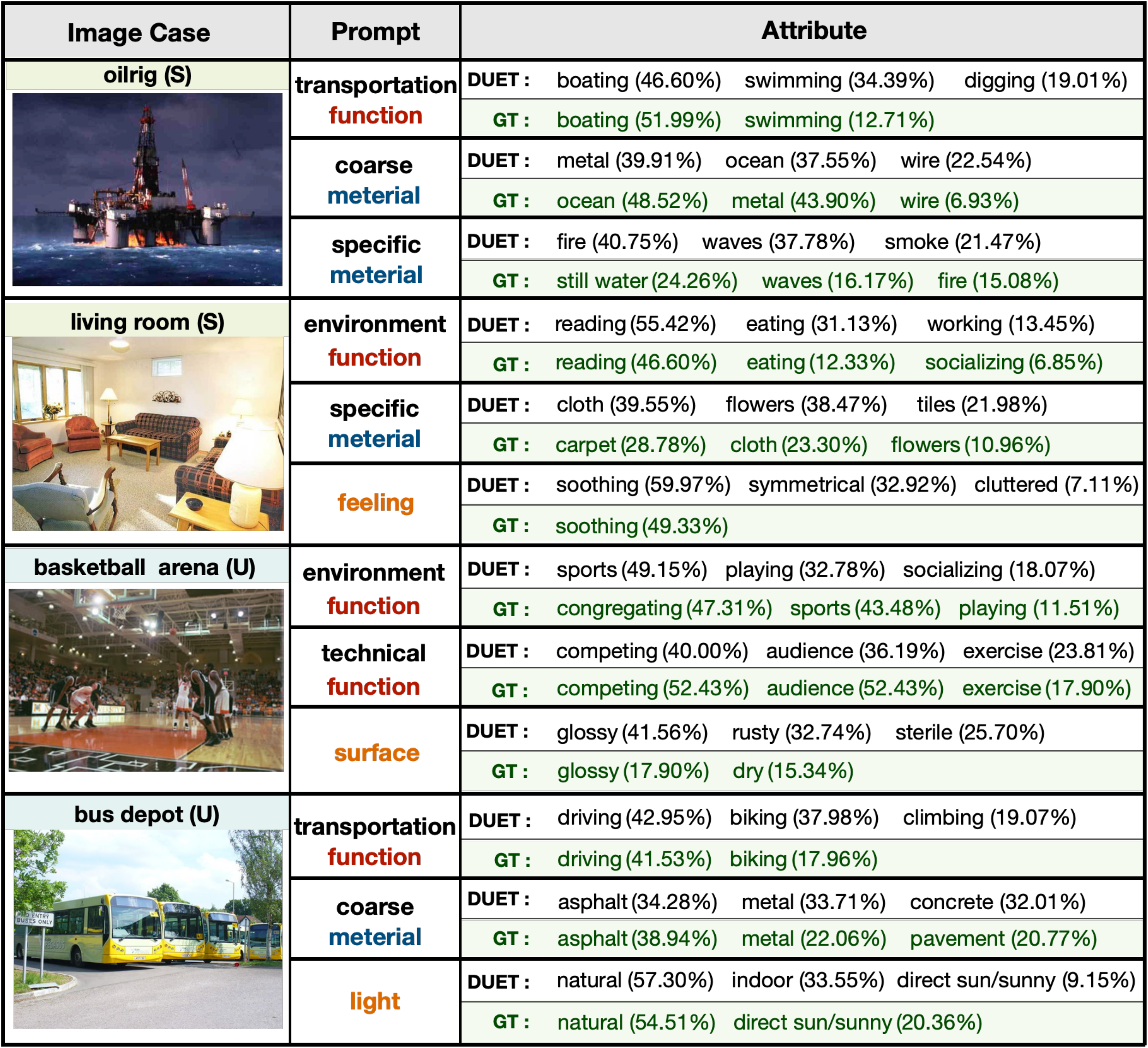}
  \vspace{-2pt}
  \caption{{Attribute prediction for interpretation.}
  %Textual attribute prediction.
  }
  \label{fig:interpretable}
  \vspace{-5pt}
\end{figure}
To proofs DUET's capability on disentangling image-specific semantic attribute\fy{s},
we feed an image into the well-trained DUET together with a crafted template: ``{\tt $...| \widehat{\mathcal{P}}$:$[MASK]|...$}'', 
where each prompt name $\widehat{\mathcal{P}}$ is involved with the {\tt [MASK]} followed to recover attribute tokens 
%(see Appendix for more details).
Figure \ref{fig:interpretable} shows the prediction results from the cases in SUN's testing set.
We observe that DUET can successfully retrieve most relevant attributes from the image given {a} concrete prompt, e.g., the ``sport'' function from a ``basketball arena'' and the ``natural'' light from an open-air ``bus depot''.

Sometimes, there also exist unreasonable attribute pairs within a class in GT attribute sets.
{For example,}
\textit{(i)} ``oilrig'' has \fy{both} ``still water'' and ``waves'' as its attributes, which is \ul{\emph{contradictory}}{;} 
\textit{(ii)} there is no ``carpet'' in this ``living room'' image, but it has high confidence.
These \fy{situations occur} when mutually exclusive attributes \fy{are} independently contained in different images within the same class, since the class-level attribute values in SUN are collected by averaging the binary labels from  annotators.
In contrast, our DUET could achieve \ul{\emph{instance-level semantic grounding}}, correctly \fy{giving} ``waves'' high score in this ``oilrig'', and ignore\fy{ing} ``carpet'' in this ``living room''.

Besides, the scarce attribute ``fire'' is confidently predicted in the ``oilrig'' image, 
\fy{while} in ``living room''\fy{,} the ``flowers'' \fy{are} recovered without ``leaves'' bound together, which demonstrate the 
%superiority 
{capability of DUET in \ul{\emph{addressing the attribute imbalance and attribute co-occurrence issues}} shown in Figure \ref{fig:case}.} 
Moreover, DUET can ground not only obvious visual attributes (e.g., ``fire'') but also those abstract properties (e.g., ``soothing'' for ``feeling'' and ``competting'' for ``technical function''), which 
{shows its potential capability}
for {knowledge inference}.

\section{Conclusion}
In this paper, we propose {an end-to-end ZSL framework named} DUET to address the well known issues of attribute imbalance and co-occurrence in zero-shot image classification. %incorporate the {semantics from a PLM into a vision transformer encoder via a self-supervised multi-modal learning paradigm}.
We design a cross-modal semantic grounding network with a novel attribute-level contrastive learning mechanism to enhance {the} model's discriminative ability toward{s} novel classes, which could well address the issues of attribute imbalance and co-occurrence in zero-shot learning.
{With extensive ablation studies and the comparison with quite a few state-of-the-art methods on four ZSL benchmarks with real-valued and binary-valued attributes, we demonstrate the effectiveness of DUET as well as its support for interpretation.}
% In the future, we might consider methods of  using different forms of  knowledge~\cite{LePa2013} for ZSL and transfer learning~\cite{CLPHC2018,LCPC2019} in multi-modal data streams.

\section*{Acknowledgement}
We want to express gratitude to the anonymous reviewers for their hard work and kind comments. This work is partially funded by NSFCU19B2027/91846204, the EPSRC project ConCur (EP/V050869/1) and the Chang Jiang Scholars
Program (J2019032).

\bibliography{aaai23}

\clearpage

\appendix
\section{Appendix}
\begin{figure*}[htbp]
  \centering
  \includegraphics[width=0.98\linewidth]{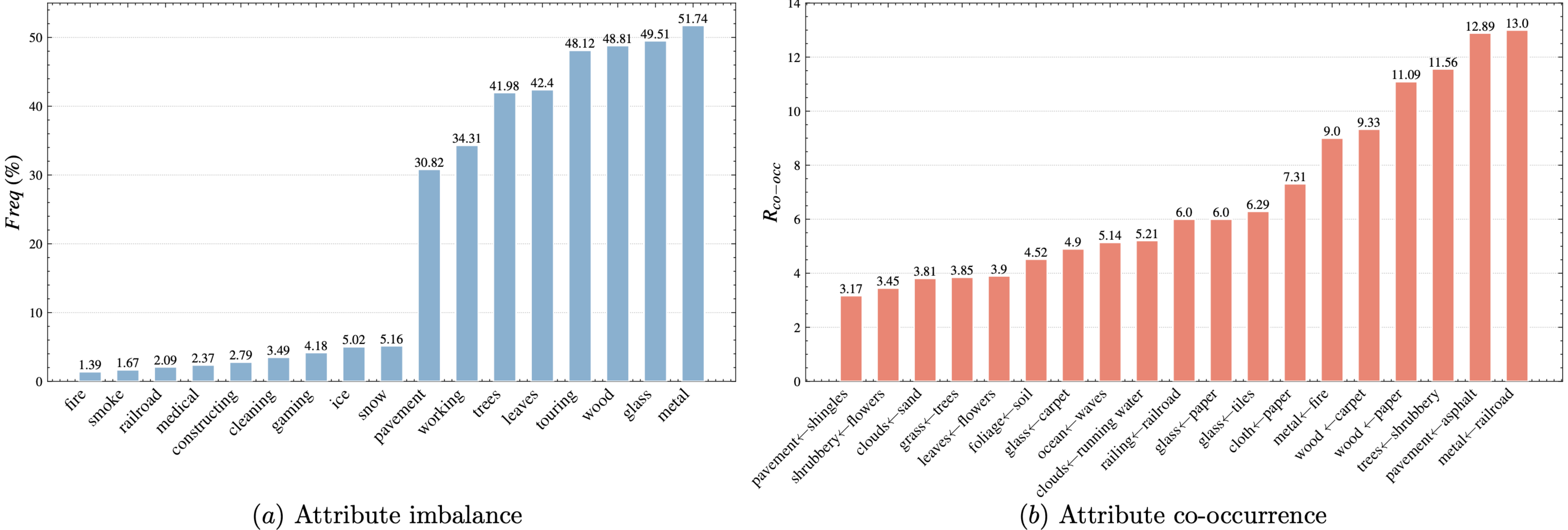}
   \vspace{-5pt}
  \caption{{Statistics about attribute co-occurrence and the imbalanced attribute distribution.}
  %Part of the statistics for representative attributes co-occurrence and imbalanced attributes  distribution.
  \vspace{-15pt}
  }
  \label{fig:imandco}
%   \vspace{-5pt}
\end{figure*}
\subsection{Prompt Details}
We cluster the attributes to define $k$ class-specific prompt set{s} for each ZSL {dataset (task)} as shown in Table \ref{tab:dataset}.

For SUN dataset, \cite{DBLP:conf/cvpr/PattersonH12}  have already divide{d} those scene attributes into four parts: ``function'', ``meterial'', ``surface\_property'', and ``spatial\_property''.
As shown in Table \ref{tab:sun}, we further split them into 14 prompt set{s} 
%$\mathcal{P}$, 
requiring that the attributes within {one set}
%the same $\mathcal{P}$
describe different characteristics toward the same aspect of the image.
{For example, we}
%e.g., 
put ``boating'' and ``driving'' from {the} original ``function'' {part} 
%type 
into {the more specific prompt set of ``transportation fuction''}.
The details {of the prompt sets of}
%for $\mathcal{P}$ in 
AWA2 \cite{DBLP:journals/pami/XianLSA19} %dataset 
are shown in Table \ref{tab:awa}.
{The attribute categories of ``color'', ``pattern'', ``shape'' and ``texture'' are from ImageNet \cite{DBLP:conf/cvpr/DengDSLL009}, and part pf the categories are from the LAD \cite{DBLP:conf/cvpr/ZhaoFLWWW19}.}

While for CUB \cite{welinder2010caltech} %dataset, 
we simply extract the prefix from the attribute names as the prompt, e.g., we extract the ``wing\_color'' prefix from the attribute`` has\_wing\_color:brown'', and make the ``brown'' as an attribute belongs to the ``wing\_color'' {prompt set.}
%$\mathcal{P}$.
Particularly, different {prompt sets}
%$\mathcal{P}$
may  have intersection{s} 
, e.g., the attribute ``spotted'' belongs to {the prompt sets of} ``wing\_pattern'' and ``head\_pattern'' 
%$\mathcal{P}$
simultaneously.
%in CUB. 
{More} details are shown in Table \ref{tab:cub}.

\subsection{Details for Experience in Sec. 4}\label{apd:hyp}

\begin{table}[htbp]
\small
\caption{{Dataset statistics.} ``$\#$ Pro.'' and ``$\#$ Att.'' {refer} to the number of prompt sets and attributes in each dataset. S/U denotes the seen/unseen class. 
}
\label{tab:dataset}
\vspace{-10pt}
\begin{center}
%  \resizebox{\linewidth}{!}{
 \scalebox{0.9}{
\begin{tabular}{l|cccc}
\specialrule{.1em}{.00em}{.00em}
Dataset  & \multicolumn{1}{c}{\# Pro.} & \multicolumn{1}{c}{\# Att.} & \multicolumn{1}{c}{\begin{tabular}[c]{@{}c@{}}\# Class \\ Total (S/U) \end{tabular}} & \multicolumn{1}{c}{\begin{tabular}[c]{@{}c@{}}\# Images \\ Training / Testing \end{tabular}} \\ 
% \specialrule{.1em}{-.15em}{.00em}
% \midrule
% \specialrule{.1em}{.00em}{.00em}
% \\ [-2ex]
\midrule
CUB      & 28                 & 312                   & 200~(150~/~50)                            & 7,057~/~4,731                            \\
SUN      & 14                 & 102                   & 717~(645~/~72)                            & 10,320~/~4,020                           \\
AWA2     & 12                 & 85                    & 50~(40~/~10)                            & 23,527~/~13795                           \\
\midrule
AWA2-KG &  17                &       120              & 50~(40~/~10)                            & 23,527~/~13795                           \\
\specialrule{.1em}{.00em}{.00em}
% \bottomrule
\end{tabular}
 }
\end{center}
\label{tab:data}
\vspace{-8pt}
\end{table}

Specifically,
% Particularly, 
we use Swin-base \cite{DBLP:conf/iccv/LiuL00W0LG21} as the vision transformer encoder for SUN, and DeiT-base \cite{DBLP:conf/icml/TouvronCDMSJ21} for CUB, AWA2 and AWA2-KG.
Both of their versions are pre-trained on ImageNet-1k.
Let $ENC_{vis}(.)$, $ENC_{lan}(.)$, $ENC_{cro}(.)$ be the encoding function of vision transformer, PLM, and cross attention layer respectively. 
We apply Bert-base \cite{DBLP:conf/naacl/DevlinCLT19} as the PLM encoder.
We take the output embedding of {\tt [CLS]} in $ENC_{vis}$ and the whole output hidden state of $ENC_{lan}$ as the input to $ENC_{cro}$.
{The cross attention layer number} $K$ is set to 1. 
For each step in $T_{CSG}$, when applying $\mathcal{L}_{acl}$, we select 2 $c^-$ (i.e., select 2 images from each $c^-$) and 1 $c^+$ 
%(i.e., select 1 image {from each $c^+$}) 
for {each}
original image $x$.
For those coefficients, we set $\lambda_{ar}$ to \hyf{0.01}, 
%$\lambda_{sc}$ to \hyf{0.001}, 
$\lambda_{con}$ to \hyf{0.05}, $r_{rap}$ to \hyf{0.5},  and $\tau$ to \hyf{0.05} for all datasets. Other dataset-specific hyperparameter{s} are shown in Table \ref{tab:hypdetail}.
We use {the} AdamW optimizer ($\beta_1 = 0.9$, $\beta_2 = 0.999$), and set the batch size to 50 {and the learning rate to 3e-5.} 
All the experiments are performed on three RTX 3090Ti GPUs.
\ul{\emph{AWA2, which has the largest scale among our benchmarks which leads to least variance, is selected for all ablation studies}}.
\begin{table}[htbp]
		\centering  
 		\small
		\caption{%Details for 
		Hyperparameter settings in four datasets.
		}
		\vspace{-5pt}
		\begin{tabular}{lccccccccc}
        \toprule
        Dataset  & $\lambda_{cmr}$ & $\lambda_{acl}$ & $\gamma$ & $\rho$ \\
        % \specialrule{.1em}{.00em}{.00em}
        \midrule
        AWA2  & 1   & 0.01   & 0.8   & 0.4    \\
        CUB   & 5   & 0.001   & 0.6   & 0.4    \\
        SUN   & 1   & 0.01   & 0.5   & 0.5    \\
        AWA2-KG & 0.5   & 0.01   & 0.7   & 0.4   \\
% \specialrule{.1em}{.00em}{.00em}	
        \bottomrule
		\end{tabular} 
% 		}
		\label{tab:hypdetail}
		\vspace{-7pt}
\end{table}

\subsection{Statistics for Attribute}
Figure \ref{fig:imandco} shows part of the statistics 
%for representative 
{about} attribute co-occurrence and {the} imbalanced attribute  distribution.
Particularly, 
we define the frequency  for those attributes as 
\begin{equation}
    Freq(a) = \frac{\sum_{{c}\prime \in \mathcal{C}} \mathbb{I}{\left[a \in \mathcal{A}^{{c}\prime}\right]}}{|\mathcal{C}|} \times 100\% \,,
\end{equation}
where $\mathbb{I}{[ a \in \mathcal{A}^{{c}\prime} ]}$ is 1 if $a$ is in $\mathcal{A}^{{c}\prime}$, and 0 otherwise.

In order to measure the degree for attribute co-occurrence, 
we define {the following metric:}
\begin{equation}
    R_{co-occ}(a_i \gets a_j) = \frac
    {\sum_{{c}\prime \in \mathcal{C}} \mathbb{I}{\left[a_i \in \mathcal{A}^{{c}\prime} ~\&~  a_j \in \mathcal{A}^{{c}\prime} \right]}}
    {\sum_{{c}\prime \in \mathcal{C}} \mathbb{I}{\left[a_i \notin \mathcal{A}^{{c}\prime} ~\&~  a_j \in \mathcal{A}^{{c}\prime} \right]}}\,.
\end{equation}
For example, ``flowers'' appears with ``leaves'' 39 times, but ``flowers'' alone only appears 10 times, then $R_{co-occ}({leaves} \gets {flowers})$ $=$ $3.9$, and $Freq({flowers})$ $=$ $49$$/$$717$$\times$$100\%$ $=$ $6.83\%$.
We observe that the distribution of different attributes varies greatly, and some unforeseen attribute pairs open appear together like (``cloth'' $\gets$ ``paper'') and (``clouds'' $\gets$``sand'').

\subsection{Semantic Grounding and Visualization}

\begin{figure*}[htbp]
  \centering
  \includegraphics[width=0.98\linewidth]{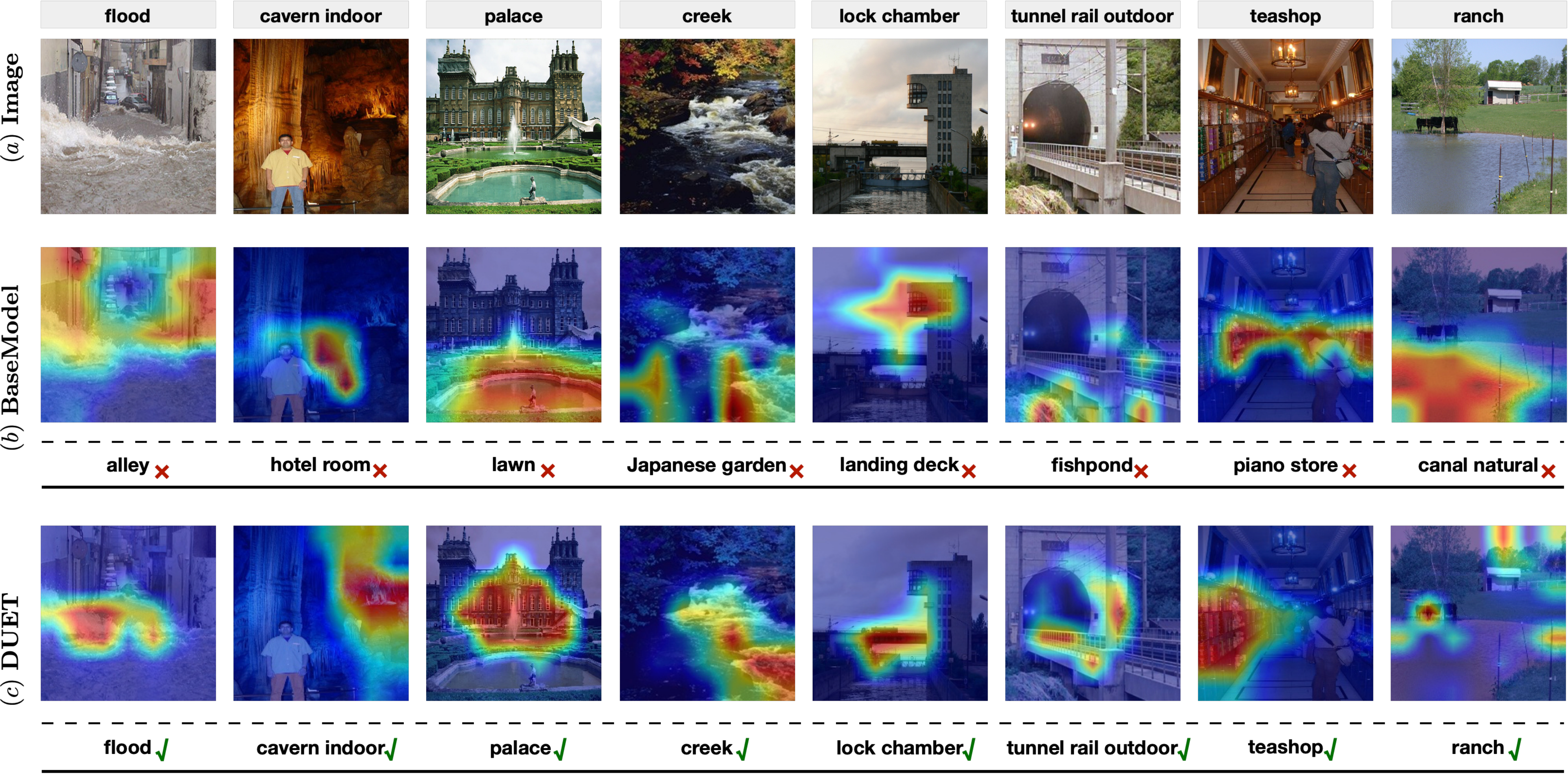}
  \caption{Visualization of attention maps together with attribute grounding results.
    (a) Original images.
  	(b) BaseModel: Only using $ENC_{vis}$ without cross-modal semantic grounding (CSG) and attribute-level contrastive learning (ACL).
	(c)DUET.
  }
  \label{fig:Visualization}
\end{figure*}

Figure \ref{fig:Visualization} visualizes the dense-attention maps on several test images {in SUN.}
%from {the} SUN dataset.
Compared with the BaseModel that only use{s} $ENC_{vis}$, our DUET can detect more accurate attribute regions that humans pay 
attention to.
{For example,}
for the ``lock chamber'' and ``tunnel rail outdoor'' images, we observe that the BaseModel fail{s} to ground the ``railroad'' attribute, 
which is an infrequent attribute {with the frequency of 2.09$\%$.}
Attribute imbalance issue may be the reason {that} BaseModel {wrongly predicts} these %scene 
{two images} as ``landing deck'' and ``fishpond''.

Moreover, 
since ``waves'',  ``soil'',  ``asphalt'' and ``running water''
always occur with ``pavement'', ``ocean'', ``foliage'' and ``clouds'' respectively, 
%we guess that the 
the BaseModel {would fail to distinguish them in predicting} 
%fail to disentangle them 
%from
the ``flood'', ``cavern indoor'', ``palace'', and ``creek'' images which {do not} contain those latter accompanying attributes.
{In contrast, DUET better  grounds those semantic attributes, leading to better overall performance.}
Lastly, our model also retains sensitivity for those key characteristics that are not obviously marked out in attribute set such as the teacups in ``teashop'' and the animals in ``ranch''.

\subsection{Sampling Strategy}
We employ a heuristic process to let the model 
%prefer to 
select those $c^+$ whose $\mathcal{A}^{c+}$ are more \ul{\emph{inconsistent}}, and  $c^-$ whose $\mathcal{A}^{c-}$ are more \ul{\emph{similar}}, compared with $\mathcal{A}^c$. 
Concretely, we measure the class similarity via 
%employing 
the Manhattan distance:
\begin{align}
	Sim(c,c\wen{^\prime}) &= 1 / Dist(c,c\wen{^\prime})\,, \\ 
\label{eq:dist}
	Dist(c,c\wen{^\prime}) &=  \sum\nolimits_{i=1}^{A} |z^c_i - z^{c\wen{^\prime}}_i |\,,
\end{align}
and then {set the}
%sample 
class $c^\prime$  as positive or negative with {a} 
%the 
probability:
\begin{align} 
\label{eq:samplen}
    P(c^- \gets c^\prime |a^c_{t}, c) &= \frac{Sim(c,c^\prime)^2}{\sum_{\hat{c} \in \mathcal{C}^-} Sim(c,\hat{c})^2 }\,,\\
\label{eq:samplep}
    P(c^+ \gets c^\prime |a^c_{t}, c) &= \frac{Dist(c,c^\prime)^2}{\sum_{\hat{c} \in \mathcal{\mathcal{C}}^+} Dist(c,\hat{c})^2 }\,,
    % Sample(c^-, c^+ |a^c_{tag}, c) = Maximize (Cover(\mathcal{A}^c-,\mathcal{A}^c)-Cover(\mathcal{A}^c+,\mathcal{A}^c))  
\end{align}
where $c^\prime \in \mathcal{C}^+$ {and} %or
$c^\prime \in \mathcal{C}^-$.

A large number of experiments were conducted and we just selected the most optimal settings (highlight with \textbf{bold}):
\begin{itemize}
\item Regarding the strategy for choosing the target attributes, we have test \textit{(i)} the random sampling; \textit{(ii)} \textbf{the linear weighted random sampling} in Eq. (\ref{eq:lwrs}); \textit{(ii)} the non-linear weighted random sampling (simple based on squaring or softmax).
\item When it comes to the strategy for sampling contrastive classes, we have tested \textit{(i)} the linear weighted random sampling;  \textit{(ii)} the non-linear weighted random sampling (\textbf{based on squaring} in Eq. (\ref{eq:samplen}) and (\ref{eq:samplep}) or simple softmax).
\item When it comes to the metric for class distance, we have tested \textit{(i)} \textbf{the Manhattan distance} in Eq. (\ref{eq:dist}); \textit{(ii)} the Euclidean distance; \textit{(iii)} the Cosine similarity.
\end{itemize}

\subsection{Cross-Attention Layer in Sec. 3.2} \label{apd:cross}
Each cross-attention layer consists of one bi-directional cross-attention block, two self-attention blocks and two feed-forward blocks. A residual connection and
layer normalization are added after each block. 
Specifically, let $h^k = \{h_1^k, ..., h_N^k\}$ be the input features of the $k$-th attention block $B_k$ ($h^0$ is the output of $Enc$), the output state $h^{k+1}$ is computed by
% \begin{equation}
\begin{align}
\tilde{h}_{i}^{k+1}=\sum\nolimits_{m=1}^{M} W_{m}^{k+1}\left\{\sum\nolimits_{j=1}^{N} A_{i, j}^{m} \cdot V_{m}^{k+1} h_{j}^{k}\right\} \,,\\
h_{i}^{l+1}=LayerNorm\left(h_{i}^{k}+\tilde{h}_{i}^{k+1}\right) \,,\\
A_{i, j}^{m} \propto \exp \left[\left(Q_{m}^{k+1} h_{i}^{k}\right)^{T}\left(K_{m}^{k+1} h_{j}^{k}\right)\right]\,,
\end{align}
% \end{equation}
where $A_{i, j}^{m}$ denotes the attention weights between elements $i$ and $j$ in the $m$-th attention head, which is normalized by $\sum_{j=1}^{N} A_{i, j}^{m}=1$. Besides, $W_m^{k+1}$, query ($Q_m^{k+1}$), key ($K_m^{k+1}$) and value ($V_m^{k+1}$) are learnable weights for $m$-th head. 
Remarkably, given any $h_i^k$,  $h_i^k$ is in the set of $\{h_j^k\}$ when $B_k$ is the self-attention block. 
Contrarily, when $B_k$ is the cross-attention block, $\{h_i^k\}$ and $\{h_j^k\}$ should be originated from different encoder{s} and have no intersection. 

\subsection{Related Work for PLMs}
Stacked by multiple transformer \cite{DBLP:conf/nips/VaswaniSPUJGKP17} blocks, the pre-trained language models (PLMs) store an amount of knowledge via learning universal language representations on Internet {corpora,}
%corpus, 
which can prevent researchers from training a new model from scratch and benefits downstream NLP tasks.
% \noindent \textbf{Mask Language Model.}
The \textbf{Mask Language Model} (MLMs) \cite{DBLP:conf/naacl/DevlinCLT19,DBLP:journals/taslp/CuiCLQY21,DBLP:conf/acl/YinNYR20} are pre-trained \fy{by} recovering the masked semantic unit in a sentence.
, \fy{which} play important roles among all those PLMs.
Specifically, the masked semantic unit in BERT \cite{DBLP:conf/naacl/DevlinCLT19} is the word token, which is a sequence of characters that are grouped together to express some basic semantic information.
Cui et al. \cite{DBLP:journals/taslp/CuiCLQY21} design a whole word masking (WWM) strategy, where the model always masks all tokens corresponding to a whole word at once. 
Yin et al. \cite{DBLP:conf/acl/YinNYR20} propose TaBERT to linearize the tables which makes PLMs understand the structured data. 

Inspired by the major success of transformer in NLP field, researchers have recently applied this architecture to CV fields \cite{DBLP:conf/iccv/WangX0FSLL0021,DBLP:conf/iclr/DosovitskiyB0WZ21,DBLP:conf/icml/TouvronCDMSJ21,DBLP:journals/corr/abs-2111-06377,DBLP:journals/corr/abs-2106-08254,DBLP:conf/iccv/LiuL00W0LG21}.
For example, the \textbf{Vision Transformer} (ViT) \cite{DBLP:conf/iclr/DosovitskiyB0WZ21} applies the sequences of image patches into pure transformer for image classification.
Based on ViT, Touvron et al. \cite{DBLP:conf/icml/TouvronCDMSJ21} 
further distill the knowledge from CNN-based model, while
Swin Transformer \cite{DBLP:conf/iccv/LiuL00W0LG21} 
computes attention within a local window via partitioning the window into multiple sub-patches. 
Previous \textbf{Vision-Language Pre-training} (VLP) models \cite{DBLP:conf/emnlp/TanB19,DBLP:conf/nips/LuBPL19,DBLP:conf/cvpr/ZhangLHY0WCG21,DBLP:conf/eccv/Li0LZHZWH0WCG20} also benefit many multi-modal tasks like visual question answering (VQA) and natural language visual reasoning (NLVR2). 
But normally,
when \fy{a multi-modal network is applied to a single-modal task, its performance drops rapidly due to the missing modality. Not to mention performing on a zero-shot single-modal task.}

In this paper, we design a novel architecture DUET, which is the first work to extend the PLM architecture to ZSL image classification task under a multi-modal paradigm.
Instead of  simply \fy{taking} a fixed semantic embedding as the input \cite{DBLP:journals/corr/abs-2201-00577}, our model leverages the knowledge within a large scale PLM via a cross-modal semantic grounding method, as shown in Figure \ref{fig:compare}(b).

\subsection{Vision Encoder Selection}
Pre-trained on ImageNet-21k, the Vision Transformer (ViT) \cite{DBLP:conf/iclr/DosovitskiyB0WZ21} applies the sequences of image patches into {a} pure transformer for image classification, where a learnable embedding is applied to the sequence of embedding patches whose state serves as the image representation.
Based on ViT, DeiT \cite{DBLP:conf/icml/TouvronCDMSJ21} 
further distill{s knowledge from a CNN-based model} 
and select {a} better hyperparameter combination, {which trains on only the ImageNet-1k database but achieves comparable performance with ViT.
Swin \cite{DBLP:conf/iccv/LiuL00W0LG21} 
use\fy{s} a multi-stage hierarchical architecture which 
computes attention within a local window via partitioning the window into multiple sub-patches.}
To capture interactions between different windows, window partitioning in Swin is gradually shifted along the {hierarchy of the network} to capture overlapping regions, 
which is suitable for capturing \ul{\emph{multi-scale attributes in SUN dataset}}.
Thus, in this paper, 
we use Swin-base as the vision transformer encoder for SUN,
and DeiT-base\footnote{\url{huggingface.co/facebook/deit-base-distilled-patch16-224}}  
for CUB, AWA2, and AWA2-KG, 
which are all pre-trained on ImageNet-1k with a fixed patch size of $16 \times 16$ pixels.
Specifically, \ul{\emph{for a fair comparison, we utilize the official conversion script}}\footnote{
\url{github.com/huggingface/transformers/blob/main/src/transformers/models/swin/convert_swin_timm_to_pytorch.py}}
\ul{\emph{to make the timm-based\footnote{\url{github.com/rwightman/pytorch-image-models}} Swin model compatible with the huggingface\footnote{\url{huggingface.co/}} transformer library}}. 

Regarding the impact of backbone on the baseline’s performance, we note that simply changing the backbone to vision transformers may not lead to better results. Concretely, we replace the backbone of a non-generative model APN \cite{DBLP:conf/nips/XuXWSA20} and a generative model ABP \cite{DBLP:conf/iccv/ZhuXLE19} from ResNet101 to our transformer backbones, and freeze (Fz) all layers for feature extraction (follow ABP) or jointly fine-tune part of the layers (follow APN) e.g. last 1(L1) or 2(L2) layers. All hyper-parameters remain unchanged for fairness. We find their performance all drop by [T1, H]\%:
\textit{(i)} \textbf{ABP}:
CUB: [10.9, 11.6]; AWA2: [9.9, 5.2]; SUN: [18.2, 14.4].
\textit{(ii)}  \textbf{APN}:
CUB: Fz: [5.6, 6.8]; L1: [4.8, 5.3]; L2: [12.2, 14.8];
AWA2: Fz: [9.9, 8.4]; L1: [6.8, 4.5];
SUN: Fz: [3.8, 6.9]; L1: [4.1, 7.6].

\subsection{Attribute prediction in Sec. 4.5}\label{apd:ap}
Specifically, we feed an image into the well-trained DUET together with a crafted template:
\begin{equation}
\small
	{\tt{..}} {\tt|} \underbrace{{\tt{color}}}_{\textrm{Prompt}} {\tt:}~  {\tt{[MASK]}}  {\tt|}  \underbrace{{\tt{has part}}}_{\textrm{Prompt}} {\tt:}~ {\tt{[MASK]}} {\tt|}	{\tt{..}}\, ,
% 	\label{eq:cell_string_representation}
\end{equation}
which contains all prompt names $\widehat{\mathcal{P}}$ with a {\tt [MASK]} token   attached behind each of them.
Those mask representations are used for attribute  token prediction, where those tokens from the attributes in {the prompt set} 
%$\mathcal{P}$ 
constitute the  predictable candidate tokens in PLM's vocabulary.
The token could be mapped into corresponding attributes (maybe more than one) for score accumulation.
Then we select those attributes whose prediction score rank within  the top-3  as the  results, and normalize  their scores 
%to sum to 1 
by dividing each of them by their {sum.}
%summation.
We also normalize those confidence score{s} from the real-value attribute vectors in {the} original SUN dataset via global Min-Max normalization, i.e. $z^c_j = 100\% \times (z^c_j - min(z_i))/(max(z_i)-min(z_i))$.
Then we choose those highly confident attributes, and denote them as the ground truth (GT)  attributes.

\subsection{Future Work}
Regarding the ways to improve performance on CUB, we could consider involving 2D relative geometry relationships via adding relative position encoding. Meanwhile, due to the intra-class image differences (images of the sample bird class in CUB are different due to e.g., different image shooting angles and different bird ages), we think DUET could be further improved by e.g., applying instance-level feature similarity as an auxiliary sample filtering strategy on the basis of our attribute-level contrastive learning (ACL) method, or pre-grounding the object, to get higher quality positive / negative samples.

Moreover, we noticed that some recent works like CoAR-ZSL\cite{DBLP:journals/corr/abs-2207-03824},  T2M-HN \cite{DBLP:journals/corr/abs-2210-15182}, VL-Taboo \cite{DBLP:journals/corr/abs-2209-06103}, and I2DFormer \cite{DBLP:journals/corr/abs-2209-10304} are proposed in the same period as ours to explore the potential of PLMs in ZSL community for zero-shot prediction/analysis. 
We insist that the trade-off between efficiency and performance is an important factor that needs to be considered in future works.

\begin{table}[ht]
		\centering  
% 		\small
		\caption{Details for {the prompts} in SUN.}
%		\vspace{-5pt}
% 		\vspace{-1mm}
		\resizebox{0.95\linewidth}{!}{
		\begin{tabular}{p{2cm}|l}
\toprule
\textbf{prompt}  & \quad \quad \quad \quad \quad \quad \quad \quad \textbf{attributes} \\
% \specialrule{.1em}{.00em}{.00em}
\midrule
 
\makecell[l]{transportation \\function}  & \makecell[l]{boating, driving, biking, climbing, swimming, \quad \quad \\ digging}   \\
\hline
\makecell[l]{environment \\function} & \makecell[l]{transporting, sunbathing, touring, hiking, \\ camping, reading, bathing, eating, socializing, \\congregating, sports, playing, working}   \\
\hline
 \makecell[l]{technical \quad \quad\quad \\function} & \makecell[l]{studying, teaching, research, diving, cleaning, \\ queuing, competing, exercise, gaming, audience, \\farming, constructing, shopping, medical, \\using tools, business, praying}   \\
 \hline
\makecell[l]{coarse \\meterial} & \makecell[l]{fencing, railing, wire, asphalt, pavement, brick,\\ concrete, metal, paper, wood, linoleum, plastic, \\ stone, ocean}   \\
\hline
 \makecell[l]{specific \quad \quad\quad \quad\\meterial} & \makecell[l]{railroad, flowers, shingles, carpet, tiles, cloth,\quad\quad \\ sand, marble, waves, running water, still water, \\ice, snow, smoke, fire}  \\
\makecell[l]{natural \\meterial} & \makecell[l]{trees, grass, vegetation, shrubbery, foliage,\\ leaves, soil, glass, clouds} \\
\hline
 light & natural light, direct sun/sunny, indoor light \\
 \hline
surface & \makecell[l]{worn, glossy, matte, sterile, damp, dry, dirty, \\rusty} \\
\hline
 temperature & warm, cold \\
 \hline
origin property & natural, man made \\
\hline
 area &open, semi enclosed, enclosed \\
 \hline
horizon &far away horizon, no horizon, rugged scene \\
\hline
 direction & vertical, horizontal \\
 \hline
feeling &  symmetrical, cluttered, scary, soothing, stressful \\

% \specialrule{.1em}{.00em}{.00em}	
\bottomrule
		\end{tabular} 
		}
		\label{tab:sun}
%		\vspace{-7pt}
\end{table}

\begin{table}[ht]
		\centering  
% 		\small
		\caption{Details for {the prompts} in AWA2.}
%		\vspace{-5pt}
% 		\vspace{-1mm}
		\resizebox{0.95\linewidth}{!}{
			\begin{tabular}{l|l}
% \toprule
% \specialrule{.1em}{.00em}{.00em}
% \multirow{2}{*}[-0.5ex]{Methods \quad } 
\toprule
\textbf{prompt}  & \quad \quad \quad \quad \quad \quad \quad \quad \quad \textbf{attributes} \\
% \specialrule{.1em}{.00em}{.00em}
\midrule
 color  & black, white, blue, brown, gray, orange, red, yellow   \\\hline
pattern   & patches, spots, stripes   \\\hline
 texture   & furry, hairless, tough skin   \\\hline
shape & big, small, bulbous, lean   \\\hline
 has part  & \makecell[l]{flippers, hands, hooves,pads, paws, long leg, long neck, \quad\quad\\ tail, chew teeth, meat teeth, buck teeth,  strain teeth, \\ horns,claws, tusks,smelly, muscle}   \\\hline
behaviour & flys, hops, swims, tunnels, walks \\\hline
 character & \makecell[l]{fast, slow, strong, weak, hibernate, inactive, nocturnal, \quad\quad\\ active, agility} \\\hline
limb & bipedal, quadrupedal \\\hline
 diet & fish, meat, plankton, vegetation, insects \\\hline
role & \makecell[l]{forager, grazer, hunter, scavenger, skimmer, stalker, \\ domestic} \\\hline
 habitat & \makecell[l]{new world, old world, arctic, coastal, desert, bush, plains, \\forest, fields, jungle, mountains, ocean, ground, water, \\ tree, cave} \\\hline
habit & hibernate, fierce, timid, smart, group, solitary, nestspot \\
% \specialrule{.1em}{.00em}{.00em}	
\bottomrule
		\end{tabular} 
		}
		\label{tab:awa}
%		\vspace{-7pt}
\end{table}

\begin{table}[ht]
		\centering  
% 		\small
		\caption{Details for {the prompts} in CUB.}
%		\vspace{-5pt}
% 		\vspace{-1mm}
		\resizebox{1\linewidth}{!}{
			\begin{tabular}{p{1.8cm}|p{6.6cm}}
% \toprule
% \specialrule{.1em}{.00em}{.00em}
% \multirow{2}{*}[-0.5ex]{Methods \quad } 
\toprule
\textbf{prompt}  & \quad \quad \quad \quad \quad \quad \quad \quad \textbf{attributes} \\
% \specialrule{.1em}{.00em}{.00em}
\midrule
 bill shape & \makecell[l]{curved, dagger, hooked, needle, hooked seabird, \\ spatulate, all purpose, cone, specialized} \\\hline
wing color & \makecell[l]{blue, brown, rainbow, purple, auburn, grey, yellow, \\olive, green, pink, orange, black, white, red, buff} \\\hline
 \makecell[l]{upperparts \\color} & \makecell[l]{blue, brown, rainbow, purple, auburn, grey, yellow, \\olive, green, pink, orange, black, white, red, buff} \\\hline
\makecell[l]{underparts \\color} & \makecell[l]{blue, brown, rainbow, purple, auburn, grey, yellow, \\olive, green, pink, orange, black, white, red, buff} \\\hline
 breast pattern & solid,spotted,striped,multi colored \\\hline
back color & \makecell[l]{blue, brown, rainbow, purple, auburn, grey, yellow, \\olive, green, pink, orange, black, white, red, buff} \\ \hline
 tail shape &  \makecell[l]{forked, rounded, notched, fan shaped, pointed, \\squared}   \\\hline
\makecell[l]{upper tail \\color} & \makecell[l]{blue, brown, rainbow, purple, auburn, grey, yellow, \\olive, green, pink, orange, black, white, red, buff} \\\hline
 head pattern & \makecell[l]{spotted, malar, crested, masked, unique, eyebrow, \\eyering, plain, eyeline, striped, capped} \\\hline
breast color & \makecell[l]{blue, brown, rainbow, purple, auburn, grey, yellow, \\olive, green, pink, orange, black, white, red, buff} \\\hline
 throat color & \makecell[l]{blue, brown, rainbow, purple, auburn,grey, yellow, \\olive, green, pink, orange, black, white, red, buff} \\ \hline
eye color & \makecell[l]{blue, brown, purple, auburn, grey, yellow,  \\olive, green, pink, orange, black, white, red, buff} \\\hline
 bill length & same as head, longer than head, shorter than head \\\hline
\makecell[l]{forehead \\color} & \makecell[l]{blue, brown, rainbow, purple, auburn, grey, yellow, \\olive, green, pink, orange, black, white, red, buff} \\\hline
 \makecell[l]{under tail \\color} & \makecell[l]{blue, brown, rainbow, purple, auburn, grey, yellow, \\olive, green, pink, orange, black, white, red, buff} \\ \hline
nape color & \makecell[l]{blue, brown, rainbow, purple, auburn, grey, yellow, \\olive, green, pink, orange, black, white, red, buff} \\\hline
 belly color & \makecell[l]{blue, brown, rainbow, purple, auburn, grey, yellow, \\olive, green, pink, orange, black, white, red, buff} \\ \hline
wing shape & round, pointed, broad, conical, long \\\hline
 size & large, small, very large, medium, very small \\ \hline
shape & \makecell[l]{water upright perching, marsh chicken, long leg, \\duck, owl, gull, hummingbird, pigeon, tree clinging, \\hawk, sandpiper, upland ground, swallow, perching} \\ \hline
 back pattern & solid, spotted, striped, multi colored \\\hline
tail pattern & solid, spotted, striped, multi colored \\\hline
 belly pattern & solid, spotted, striped, multi colored \\ \hline
\makecell[l]{primary \\color} & \makecell[l]{blue, brown, rainbow, purple, auburn, grey, yellow, \\olive, green, pink, orange, black, white, red, buff} \\\hline
 leg color & \makecell[l]{blue, brown, rainbow, purple, auburn, grey, yellow, \\olive, green, pink, orange, black, white, red, buff} \\ \hline
bill color & \makecell[l]{blue, brown, rainbow, purple, auburn, grey, yellow, \\olive, green, pink, orange, black, white, red, buff}  \\\hline
 crown color & \makecell[l]{blue, brown, rainbow, purple, auburn, grey, yellow, \\olive, green, pink, orange, black, white, red, buff} \\\hline
wing pattern & solid, spotted, striped, multi colored \\

% \specialrule{.1em}{.00em}{.00em}	
\bottomrule
		\end{tabular} 
		}
		\label{tab:cub}
%		\vspace{-7pt}
\end{table}

\end{document}